%% file: camera-ready.tex
\documentclass[sigconf, screen]{acmart}
\AtBeginDocument{%
  }

\copyrightyear{2025}
\acmYear{2025}
\setcopyright{acmlicensed}
\acmConference[MM '25]{Proceedings of the 33rd ACM International Conference on Multimedia}{October 27--31, 2025}{Dublin, Ireland}
\acmBooktitle{Proceedings of the 33rd ACM International Conference on Multimedia (MM '25), October 27--31, 2025, Dublin, Ireland}
\acmDOI{10.1145/3746027.3754856}
\acmISBN{979-8-4007-2035-2/2025/10}

\usepackage{balance}
\usepackage{multirow}
\usepackage{bbding}
\usepackage{pifont}
\usepackage{threeparttable}

\input{package}
\input{preamble}
\usepackage{tabularx}
\definecolor{dgreen}{RGB}{20,139,101}
\usepackage{tcolorbox}
\usepackage{ragged2e}

\newcommand*\colourcheck[1]{%
  \expandafter\newcommand\csname #1check\endcsname{\textcolor{#1}{\ding{52}}}%
}
\colourcheck{mygreen}
\colourcheck{myred}
\usepackage{subfig}  
\usepackage{color, colortbl}
\definecolor{Gray}{gray}{0.85}
\definecolor{mygreen}{rgb}{0.0, 0.47, 0.0}
\definecolor{myred}{rgb}{0.85, 0.0, 0.0}
\definecolor{myblue}{rgb}{0.0, 0.0, 0.85}
\definecolor{myorange}{rgb}{0.85, 0.45, 0.1}
\definecolor{deepcarrotorange}{rgb}{0.91, 0.41, 0.17}
\definecolor{amber}{rgb}{0.95, 0.60, 0.0}
\definecolor{amaranth}{rgb}{0.9, 0.17, 0.31}

\begin{document}


\title[Benchmarking and Bridging Emotion Conflicts for Multimodal Emotion Reasoning]{Benchmarking and Bridging Emotion Conflicts for \\ Multimodal Emotion Reasoning}
%
\author{Zhiyuan Han}
\affiliation{%
  \institution{University of Science and Technology of China}
    \city{Heifei}
  \country{China}
}
\email{aaronhan@mail.ustc.edu.cn}

\author{Beier Zhu$^\dag$}
\affiliation{%
  \institution{Nanyang Technological University}
  \country{Singapore}
  }
\email{beier.zhu@ntu.edu.sg}

\author{Yanlong Xu}
\affiliation{%
  \institution{University of Science and Technology of China}
    \city{Heifei}
  \country{China}
}
\email{kc30@mail.ustc.edu.cn}

\author{Peipei Song}
\affiliation{%
  \institution{University of Science and Technology of China}
    \city{Heifei}
  \country{China}
}
\email{songpeipei@ustc.edu.cn}

\author{Xun Yang$^\dag$}
\affiliation{%
  \institution{University of Science and Technology of China}
    \city{Heifei}
  \country{China}
}
\email{xyang21@ustc.edu.cn}

\renewcommand{\shortauthors}{Zhiyuan Han, Beier Zhu, Yanlong Xu, Peipei Song, and Xun Yang}





\input{define}

\input{section/0_abstract}
\input{section/1_introduction}

\input{section/2_relatedworks}

\input{section/3_dataset}

\input{section/4_analyses}

\input{section/5_method}

\input{section/6_expriments}

\input{section/7_conclusion}

\bibliographystyle{ACM-Reference-Format}
\balance
\bibliography{sample-base}

\input{section/X_appendix}

\end{document}

%% file: package.tex
\usepackage{enumitem}
\usepackage{dsfont}

\usepackage{algorithm}
\usepackage{algpseudocode}
\usepackage[export]{adjustbox}

%% file: preamble.tex
\usepackage{color}


%% file: define.tex
\newcommand{\zbe}[1]{{\color{blue}#1}}

\def\eg{\emph{e.g.}} 
\def\Eg{\emph{E.g}}
\def\ie{\emph{i.e.}} 
\def\Ie{\emph{I.e}}
\def\cf{\emph{cf} } 
\def\Cf{\emph{Cf}}
\def\etc{\emph{etc}} 
\def\vs{\emph{vs}}
\def\wrt{w.r.t. } 
\def\dof{d.o.f}
\def\iid{i.i.d} 
\def\wolog{w.l.o.g}
\def\etal{\emph{et al}}

\def\ourdataset{\texttt{CA-MER}} 
\def\ourmethod{\texttt{MoSEAR}}
\def\ourtrain{\texttt{MoSE}}
\def\ourinfer{\texttt{AR}}

\newcommand{\tableCellHeight}{1}
\newcommand{\tabstyle}[1]{
  \setlength{\tabcolsep}{#1}
  \renewcommand{\arraystretch}{\tableCellHeight}
  \centering
  \small
}

%% file: section/0_abstract.tex
\newcommand{\unnumberedfootnote}[1]{%
  \begingroup
  \renewcommand{\thefootnote}{}%
  \footnote{#1}%
  \addtocounter{footnote}{-1}%
  \endgroup
}

\begin{abstract}
Despite their strong performance in multimodal emotion reasoning, existing Multimodal Large Language Models (MLLMs) often overlook the scenarios involving emotion conflicts, where emotional cues from different modalities are inconsistent.
To fill this gap, we first introduce \ourdataset, a new benchmark designed to examine MLLMs under realistic emotion conflicts. It consists of three subsets: video-aligned, audio-aligned, and consistent, where only one or all modalities reflect the true emotion. However, evaluations on our \ourdataset~reveal that current state-of-the-art emotion MLLMs systematically over-rely on audio signal during emotion conflicts, neglecting critical cues from visual modality. 
To mitigate this bias, we propose \ourmethod, a parameter-efficient framework that promotes balanced modality integration. \ourmethod~consists of two modules:
(1) \ourtrain, modality-specific experts with a regularized gating mechanism that reduces modality bias in the fine-tuning heads; and
(2) \ourinfer, an attention reallocation mechanism that rebalances modality contributions in frozen backbones during inference. 
Our framework offers two key advantages: it mitigates emotion conflicts and  improves performance on consistent samples—without incurring a trade-off between audio and visual modalities. 
Experiments on multiple benchmarks—including MER2023, EMER, DFEW, and our \ourdataset—demonstrate that \ourmethod~achieves state-of-the-art performance, particularly under modality conflict conditions.
The code is available at \url{https://github.com/ZhiyuanHan-Aaron/MoSEAR}
\unnumberedfootnote{$^\dag$Corresponding authors}
\end{abstract}

\ccsdesc[300]{Computing methodologies~Activity recognition and understanding}

\keywords{Explainable Multimodal Emotion Reasoning, Multimodal Large Language Model, Multimodal Emotion Conflicts, Modality Bias}
\maketitle

%% file: section/1_introduction.tex
\section{Introduction}
\label{introduction}

Understanding human emotions is essential for effective human-computer interaction, enabling applications such as educational assistance~\cite{imani2019survey} and psychological counseling~\cite{hutchison2017emotion}. Early emotion recognition methods typically focus on single-modality inputs~\cite{devlin2018bert, lei2023instructerc}, rely on closed-set emotion categories~\cite{fan2021lssed,jiang2020dfew, liu2022mafw} , and lack explanatory reasoning~\cite{lian2023mer,yu2020ch,cao2014crema}. Recently, Multimodal Large Language Models (MLLMs)~\cite{hurst2024gpt,liu2024deepseek}  have emerged as powerful tools capable of processing and reasoning across multimodal information (\eg, video, audio, and text), enabling open-set emotion recognition and interpretable predictions~\cite{lian2023explainable,cheng2024emotion,yang2025omni}.

\begin{figure*}[t]
\centering
\includegraphics[width=\linewidth]{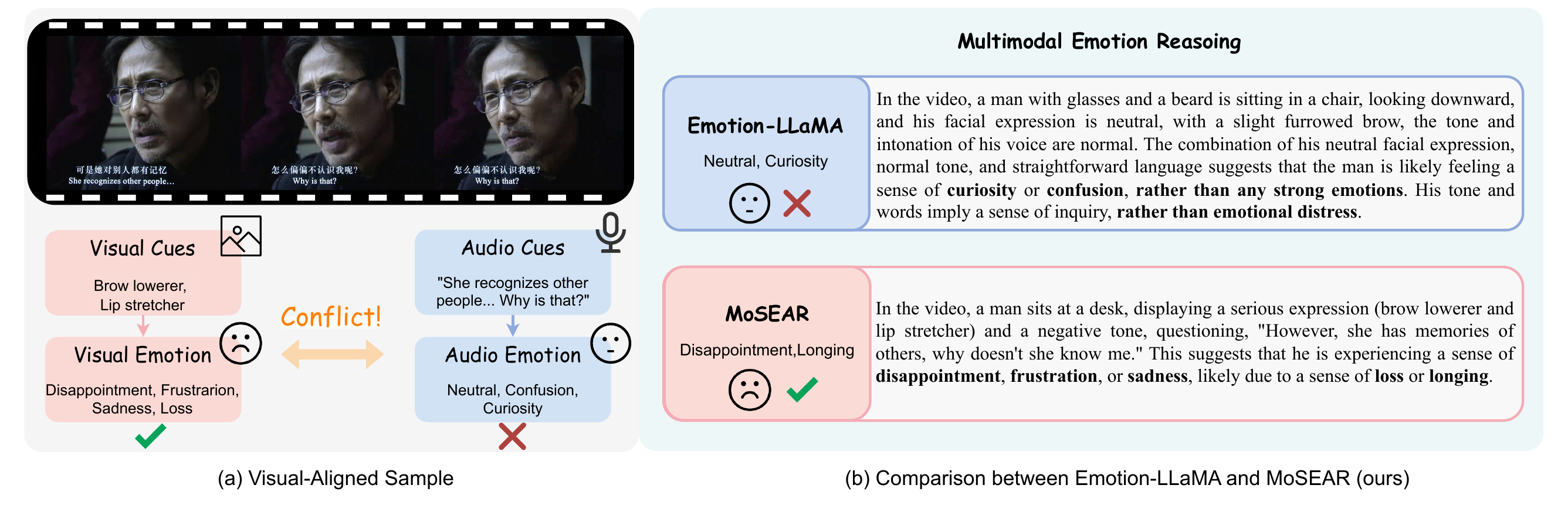}
\captionsetup{skip=6pt} 
\caption{Example of an emotion conflict case with reasoning outputs from Emotion-LLaMA and our \ourmethod.
(a) A visual-aligned sample in which the character’s facial expression conveys a clear sense of disappointment.
(b) Our \ourmethod~ provides a correct emotion reasoning, while Emotion-LLaMA produces an incorrect one under emotion conflict.
}
\label{fig:fig1}
\end{figure*}

Despite promising advances, existing emotion MLLMs and multimodal emotion benchmarks often overlook or intentionally avoid scenarios involving \textbf{multimodal emotion conflicts}~\cite{yang2025omni,lian2024affectgpt}. For instance, Omni-Emotion~\cite{yang2025omni} explicitly discards emotionally inconsistent samples. 
This limitation is problematic because humans naturally express emotions inconsistently across different modalities due to social norms, emotion regulation, or unconscious emotional leakage~\cite{bargh2007nonconscious,gross2014emotion}. 
As shown in Figure~\ref{fig:fig1}(a), an individual's disappointed and sad facial expression contrasts with their deliberately composed neutral tone.\footnote{Background: The man’s beloved wife is suffering from amnesia and no longer recognizes him. Despite his calm tone, his facial expression reveals sorrow and suppression.}

To investigate how MLLMs handle emotion conflicts, we first introduce \texttt{C}onflict-\texttt{A}ware \texttt{M}ultimodal \texttt{E}motion \texttt{R}easoning (\textbf{\ourdataset}) dataset, a new benchmark comprising three subsets, \ie, video-aligned, audio-aligned, and consistent.
Specifically, the video-aligned and audio-aligned subsets contain samples where only one modality (either video or audio) matches the true emotion, while the other modalities conflict; the consistent subset includes samples where all modalities uniformly express the true emotion. Through extensive evaluation on this benchmark, we reveal that existing MLLMs exhibit systematic \textbf{over-reliance on audio modality} in emotion conflicts, neglecting critical cues from visual modalities. Specifically, we observe a substantial performance drop in the video-aligned subset, \eg, Emotion-LLaMA~\cite{cheng2024emotion}, the current SoTA, achieves $12\%$ lower performance on the video-aligned subset than on the audio-aligned subset (Sec.~\ref{sec:sota_comparison}).
Figure~\ref{fig:fig1}(b) illustrates a concrete example where Emotion-LLaMA overly relies on acoustic cues in emotion conflicts, disregarding visual cues that humans can easily interpret as the true emotion.
This finding is further supported by attention analysis, which reveals that intermediate model layers attend more to audio tokens than to visual ones (Sec.\ref{sec:audio_bias}). Such audio bias can be attributed to the extreme imbalance between video and audio token number, as supported by our empirical evidence (Sec.\ref{sec:audio_bias}).

To address the issue, we propose \texttt{Mo}dality-\texttt{S}pecific \texttt{E}xperts and \texttt{A}ttention \texttt{R}eallocation (\textbf{\ourmethod}), a framework that mitigates modality bias during emotion conflicts by explicitly encouraging balanced modality integration. 
Specifically, \ourmethod~consists of two complementary modules: (1) \texttt{Mo}dality-\texttt{S}pecific \texttt{E}xperts (\textbf{\ourtrain}) to address bias in \textit{fine-tuning heads}, and (2) \texttt{A}ttention \texttt{R}eallocation (\textbf{\ourinfer}) to reduce bias in \textit{frozen backbones}. 
Given a pre-trained MLLM, we design \ourtrain~ — parameter-efficient modules, each aimed at enhancing feature representation across different modalities. 
Different from previous modality fusion methods~\cite{wu2024omni,luo2024mono,lin2024moma,wang2023image,shen2023scaling}, our \ourtrain~implements a regularized gating mechanism that introspects the importance of visual and non-visual information, preventing over-reliance on any single modality.
During inference, our \ourinfer~performs sample-wise attention re-balancing in frozen backbones when excessive focus on a specific modality is detected. Note that, unlike previous attention-shifting methods~\cite{liu2024paying}, our \ourinfer~does not trade off performance between visual and audio modalities: gains on video-aligned test data do not compromise audio-aligned performance. 
Moreover, our method improves performance on the emotion consistent subset, demonstrating its effectiveness beyond conflict scenarios.
We will show the evidence in Sec.~\ref{sec:ablation}.

We evaluate our \ourmethod~on multimodal emotion recognition and reasoning tasks across multiple datasets, including our \ourdataset, MER2023\cite{lian2023mer}, EMER\cite{lian2023explainable}, and DFEW\cite{jiang2020dfew}. Experimental results show that \ourmethod~consistently achieves state-of-the-art performance, especially on the three subsets of our \ourdataset~benchmark. Our contributions in this paper  include:

\begin{figure*}[t]
\centering
\includegraphics[width=\linewidth]{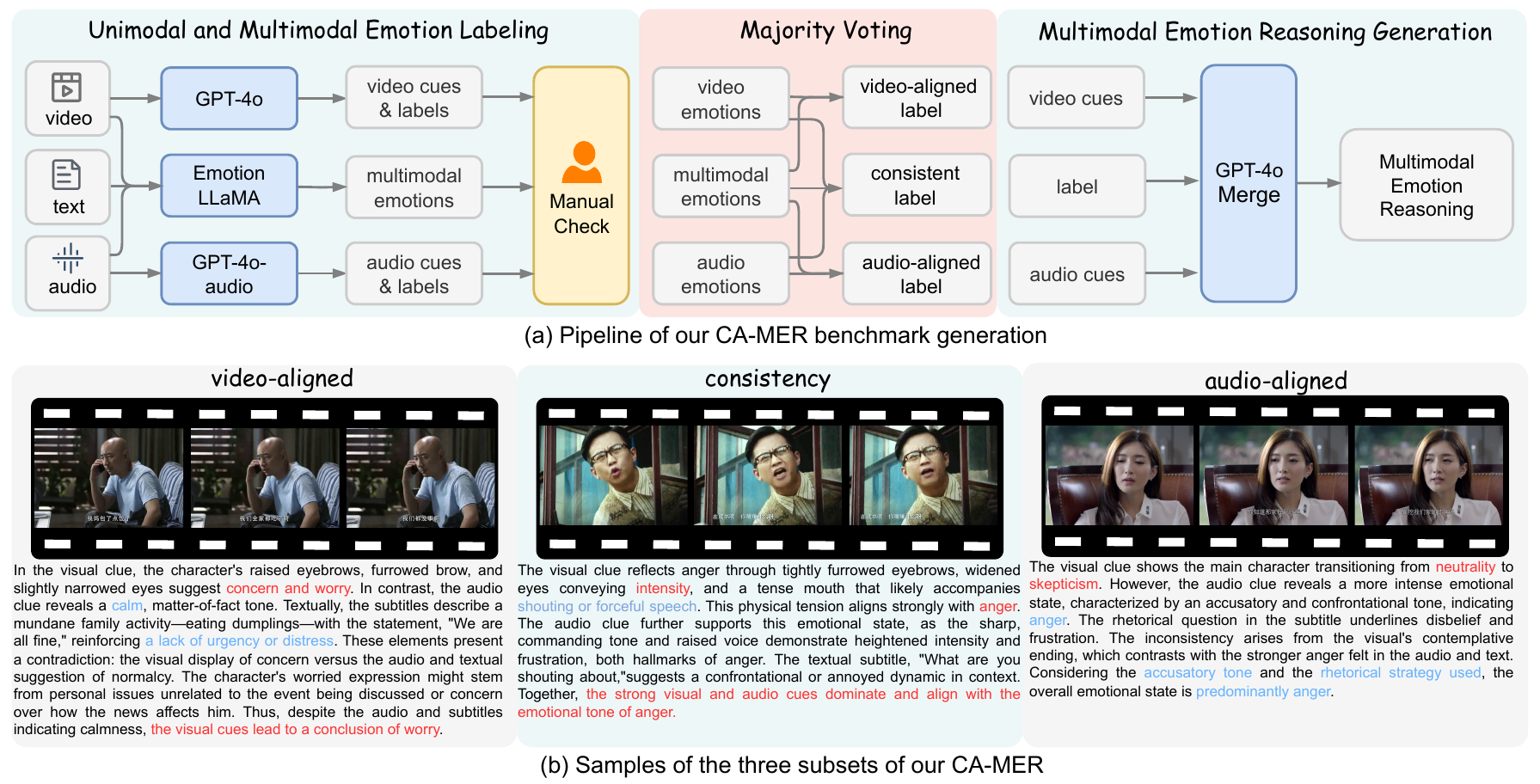}
\captionsetup{skip=6pt} 
\caption{ of \ourdataset~ construction.
(a) The three-stage construction process of our \ourdataset~ dataset.
(b) Example samples from the three subsets: video-aligned, consistent, and audio-aligned. Video cues are in {\color{myred}red}, and audio cues are in {\color{myblue}blue}.}
\label{fig:dataset}
\end{figure*}
\begin{itemize}[leftmargin=1em] 
    \item \textbf{Benchmark:} We introduce {\ourdataset}, a novel multimodal emotion reasoning benchmark comprising video-aligned, audio-aligned, and consistent subsets, enabling the evaluation of MLLMs under realistic emotion conflict scenarios.
    \item \textbf{Findings:} We identify and analyze the systematic over-reliance of existing MLLMs on the audio modality in emotion conflicts. Our empirical analysis confirms that a key factor contributing to this modality bias is the extreme imbalance in token counts between audio and video modalities.
    \item \textbf{Methodology:} We propose {\ourmethod}, a framework that addresses modality bias during emotion conflicts by integrating two modules: {\ourtrain}, which reduces bias in fine-tuning heads, and {\ourinfer}, which reallocates attention in frozen backbones without compromising modality performance.
    \item \textbf{Performance:} Experimental results demonstrate that {\ourmethod}  achieves state-of-the-art performance across multiple datasets, with notable improvements on the challenging {\ourdataset}.
\end{itemize}

%% file: section/2_relatedworks.tex
\section{Related Work}
\label{sec:related_work}

\noindent \textbf{Multimodal large language models.} 
The recent rapid development of large language models (LLMs)~\cite{pan2023finding,pan2025precise,hu2024psycollm}  has led to numerous efforts incorporating multimodal information~\cite{yang2021deconfounded,yang2022video,yang2024robust,zhou2025egotextvqa,yang2024learning} into LLMs, resulting in the emergence of multimodal large language models (MLLMs)~\cite{alayrac2022flamingo, qwen, shikra, vicuna2023, peng2023kosmos, wang2023visionllm,zhao2025unsupervised} . It have attracted significant attention for their remarkable ability to reason across diverse modalities. These models can be categorized according to the modalities they are designed to process. For example, LLaVA~\cite{liu2024llava} and GPT-4V~\cite{openai2023gpt4v} specialize in image-text understanding; Video-Chat~\cite{maaz2023video}, Chat-UniVi~\cite{jin2024chat}, and mPlug-Owl3~\cite{ye2024mplug} are tailored for video-text interactions; SALMONN~\cite{tang2023salmonn} and Qwen-Audio~\cite{chu2023qwen} excel in audio understanding; GPT-4o~\cite{hurst2024gpt} and ViTA1.5~\cite{fu2025vita} can process audio, video, and text. Although these models possess general reasoning capabilities, accurate multimodal emotion analysis still demands domain-specific knowledge.

\noindent \textbf{Multimodal emotion recognition and reasoning.}  Early works primarily focus on emotional video captioning~\cite{song2024emotional,song2023emotion,song2023contextual} and multimodal emotion recognition, such as MER 2023~\cite{lian2023mer} and DFEW~\cite{jiang2020dfew}, which classify emotions within a fixed label space. Recently, there has been growing interest in leveraging MLLMs for complex multimodal emotion reasoning tasks~\cite{lian2024mer,lian2023explainable,lian2024affectgpt,cheng2024emotion,yang2025omni,zhao2025humanomni,zhao2025r1}. Unlike traditional emotion recognition, these reasoning tasks generate predictions in an open-vocabulary manner accompanied by corresponding explanation. For instance, EMER~\cite{lian2023explainable} introduces an explainable multimodal emotion reasoning benchmark and leverages text generation to provide step-by-step reasoning. EmoVIT~\cite{xie2024emovit} combines visual cues with instruction tuning but ignores audio information. AffectGPT~\cite{lian2024affectgpt} was trained on the EMER task, but its limited training scale reduced its generalization ability.  Emotion-LLaMA~\cite{cheng2024emotion} and Omni-Emotion~\cite{yang2025omni}  introduce novel emotion reasoning datasets and build corresponding models. However, current emotion MLLMs and emotion reasoning benchmarks overlook the commonly encountered  emotion conflict phenomenon. In this paper, we introduce a novel dataset, \ourdataset, to evaluate this phenomenon and reveal that current MLLMs still struggle with it. This underscores the need for our proposed \ourmethod, which excels in handling emotion conflicts by mitigating modality bias.

\noindent\textbf{Attention-based intervention.} 
Attention-based approaches~\cite{liu2024paying,kang2025see,jiang2024devils} have been explored as training-free techniques to mitigate hallucinations in large vision-language models—namely, the generation of objects or relations absent from the visual input~\cite{ji2023survey,rawte2023survey,liu2024survey}.
However, these prior methods often intervene in attention in a coarse-grained manner. For example, PAI~\cite{liu2024paying} treats the visual attention of all layers indiscriminately, proportionally amplifying the attention weights assigned to visual tokens. Devils~\cite{jiang2024devils} first identifies which LLM layers require intervention by analyzing attention patterns, yet still intervenes in every attention head within these layers without distinction. However, these methods encounter a trade-off between the audio and visual modalities in the multimodal emotion reasoning task. In contrast, our \ourinfer~first locates the biased layers and heads with fine granularity, then adjusts the attention while preserving the overall distribution structure of attention weights. This approach avoids inter-modal trade-offs and achieves performance improvements across all scenarios.


%% file: section/3_dataset.tex
\section{Conflict-Aware Multimodal Emotion Reasoning Benchmark}
\label{sec:dataset}

\input{figs/attn_analysis}

Multimodal emotion conflicts are common, as humans often express emotions inconsistently across modalities due to social norms, emotion regulation, or unconscious leakage~\cite{bargh2007nonconscious,gross2014emotion}.
However, there is a shortage of multimodal emotion datasets for evaluating MLLMs in emotion conflicts. To fill this gap, we curate the Conflict-Aware Multimodal Emotion Reasoning dataset (\ourdataset), which comprises three subsets: video-aligned, audio-aligned, and consistent. The video- and audio-aligned subsets comprise samples where the respective modality reflects the true emotion, while the others present conflicting cues. The consistent subset includes samples that both modalities express the true emotion. We build our \ourdataset~based on MER~\cite{lian2023mer}, a widely used multimodal emotion dataset featuring annotated TV/movie clips with visual, audio, and textual cues. Figure~\ref{fig:dataset}(a) presents the three-stage pipeline for dataset construction.

\noindent \textbf{Stage 1: unimodal and multimodal emotion labeling.} For unimodal labeling, we use GPT-4o~\cite{hurst2024gpt} to independently process audio and visual inputs, generating modality-specific emotion descriptions, which are then categorized into one of nine emotion classes: \textit{\{angry, happy, surprise, fear, sad, worry, neutral, doubt, contempt\}}. For multimodal labeling, we employ Emotion-LLaMA~\cite{cheng2024emotion} to predict an emotion label from the same label set based on combined multimodal inputs. Note that all labels are \textit{manually verified by three annotators} to prevent erroneous predictions.

\noindent \textbf{Stage 2: majority voting.} 
We perform majority voting over the three labels (audio, visual, and multimodal) to determine the final emotion label. Based on the agreement among the labels, each sample is  assigned to one of the three subsets: {(1) video-aligned}: video and multimodal labels agree, but the audio label differs. {(2) audio-aligned}:  audio and multimodal labels agree, but the video label differs. (3) consistent: all three labels agree. In addition, samples with fully inconsistent labels are discarded.

\noindent \textbf{Stage 3: multimodal emotion reasoning generation.} We input the visual and audio emotion descriptions from Stage 1, together with the emotion label of Stage 2, into GPT-4o to generate the final multimodal emotion reasoning process.
Finally, we construct \ourdataset, comprising 1500 evaluation samples, with 500 samples in each subset. Figure~\ref{fig:dataset}(b) illustrates samples from our \ourdataset.

%% file: figs/attn_analysis.tex
\begin{figure*}[t!]
\centering
\includegraphics[width=\linewidth]{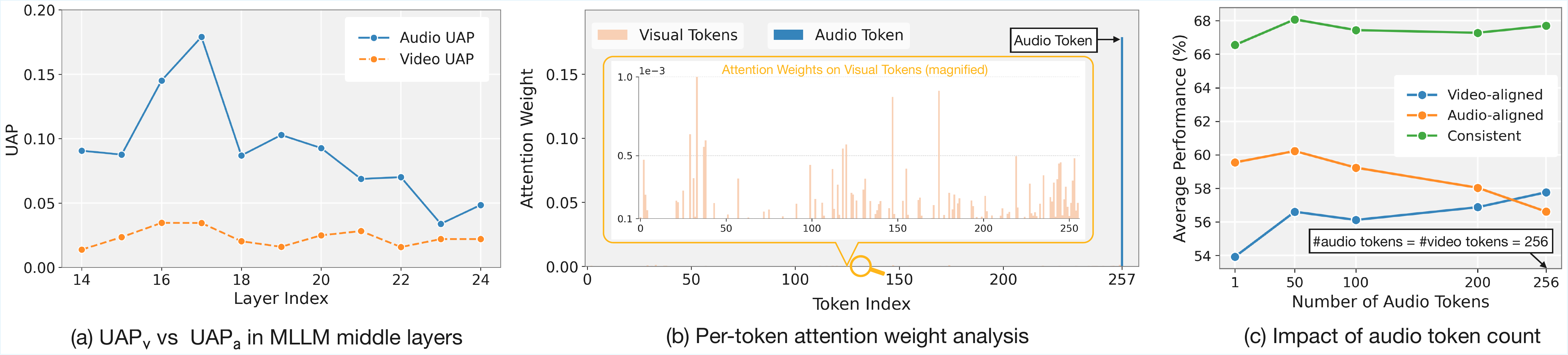}
\captionsetup{skip=6pt} 
\caption{Analyses of modality bias in emotion conflicts}
\label{fig:understanding}
\end{figure*}

%% file: section/4_analyses.tex
\section{Understanding MLLM Reasoning in Emotion Conflicts}
\label{sec:audio_bias}

Extensive evaluation on our \ourdataset~benchmark reveals that current emotion MLLMs (\eg, SALMONN\cite{tang2023salmonn}, ViTA1.5~\cite{fu2025vita}, and Emotion-LLaMA~\cite{cheng2024emotion}) perform significantly worse on the video-aligned subset than on the audio-aligned subset (see Table \ref{tab:CA-MER}). This indicates an over-reliance on acoustic cues in the presence of emotion conflicts, with insufficient attention to visual information during reasoning. 
In this section, we further investigate the phenomenon by analyzing MLLMs’ attention patterns and attributing the observed \emph{audio bias} to the extreme imbalance between video and audio token counts.

\noindent\textbf{Attention Pattern Analysis}. Analyzing attention patterns is a widely used approach to understanding the internal behavior of MLLMs~\cite{jiang2024devils,kang2025see,liu2024paying}. 
We begin by introducing our analytical metric: 
{Unimodal Attention Proportion (UAP)}, which quantifies the proportion of attention assigned to each modality. 
Let $L$ be the number of Transformer layers in MLLMs, each with $H$ attention heads. For layer $\ell$, we denote the $m$ visual tokens as $\mathcal{V} = \{\mathbf{v}_1, \dots, \mathbf{v}_m\}$ and the $n$ audio tokens as $\mathcal{A} = \{\mathbf{a}_1, \dots, \mathbf{a}_n\}$. The MLLM generates responses in an autoregressive manner. At decoding step $k$, let $\mathbf{y}_k$ be the generated token, and $\omega_{h}(\mathbf{x})$ denote its attention weight on a previous token $\mathbf{x}$ in head $h \in [H]$.
Without loss of generality, we assume that $\mathbf{y}_k$ is the first response token that reflects the emotion. The unimodal attention proportion for the visual and audio modalities at layer $\ell$ is defined as:
\begin{equation}
\text{UAP}_\mathsf{v}=\frac{1}{H}\sum_{\mathbf{v}\in \mathcal{V}}\omega_{h}(\mathbf{v}), \quad \text{UAP}_\mathsf{a}=\frac{1}{H}\sum_{\mathbf{a}\in \mathcal{A}}\omega_{h}(\mathbf{a})
\end{equation}
$\text{UAP}$ quantifies the dependence of the token $\mathbf{y}_k$ on each modality: a higher $\text{UAP}_\mathsf{v}$ (or $\text{UAP}_\mathsf{a}$) indicating \textit{a greater contribution from visual (or audio) tokens} during the generation of $\mathbf{y}_k$.

Building on the findings of~\cite{jiang2024devils} that MLLMs primarily integrate visual information in the middle layers, we center our analysis on these layers.
Specifically, we compute the average $\text{UAP}_\mathsf{v}$ and $\text{UAP}_\mathsf{a}$ across the middle layers for failure cases in the video-aligned subset, using Emotion-LLaMA~\cite{cheng2024emotion}. The results, shown in Figure~\ref{fig:understanding}(a), illustrate that the intermediate layers of the model place significantly more attention on audio tokens than on preceding visual tokens, even when the visual modality conveys the true emotion. 
In addition, we compute the per-token attention weights by averaging across the middle layers and visualize them in Figure~\ref{fig:understanding}(b). The results show that attention to audio tokens is significantly higher, while attention to visual tokens is sparse and minimal—for example, attention weights on audio tokens exceed 0.15, whereas the maximum weight on visual tokens is only around $10^{-3}$. These observations confirm the audio bias of MLLM in emotion conflicts. 

\noindent\textbf{Key factor: video-audio token imbalance.} We find that one key factor contributing to the systematic audio bias in MLLMs is the {extreme imbalance between the number of video and audio tokens.} 
We observe a significant disparity in token counts, with video tokens outnumbering audio tokens by at least an order of magnitude. For example, Emotion-LLaMA~\cite{cheng2024emotion} uses 256 video tokens but only 1 audio token; M2-Omni~\cite{guo2025m2} allocates 6144 tokens to video and 256 to audio; and for an 8-second sample, ViTA1.5~\cite{fu2025vita} processes 2048 visual tokens versus 93 audio tokens. 
Due to its high dimensionality, video information tends to be sparse and noisy, causing MLLMs to favor compact audio cues for reasoning. To support this hypothesis, we train a series of models based on Emotion-LLaMA by progressively duplicating audio tokens until their count matches that of video tokens (see Appendix C.2 for the training details). Note that this operation \textit{does not introduce extra audio information}—it simply replicates existing audio tokens to balance the modality sizes. 

In Figure~\ref{fig:understanding}(c), we evaluate these models on \ourdataset~and note several key observations:
\textbf{(1)} Increasing the number of audio tokens improves performance on the video-aligned subset ({\color{myblue}blue line}) but degrades it on the audio-aligned subset ({\color{myorange}orange line}), revealing a trade-off driven by token imbalance.
\textbf{(2)} When audio and video tokens are equal (\ie, 256), performance on the video-aligned subset surpasses the audio-aligned one, indicating a reversed bias toward the visual modality.
\textbf{(3)} The consistent subset shows no significant change due to the trade-off between the audio and video modalities ({\color{mygreen}green line}). 
While token imbalance is a key factor behind modality bias, simply increasing audio tokens introduces higher inference costs without truly addressing the root cause due to its trade-off nature~\cite{NEURIPS2024_8cc7e150,zhu2022cross,zhu2023generalized}. Instead, we propose \ourmethod~in the next section—a more effective solution that mitigates modality bias and improves performance on consistent samples.

%% file: section/5_method.tex
\section{Methods}
\label{methods}

\begin{figure}[t]
\centering
\includegraphics[width=\linewidth]{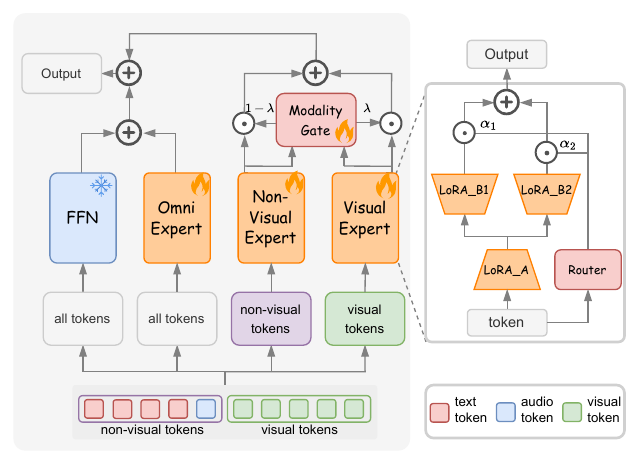}
\captionsetup{skip=6pt} 
\caption{Illustration of modality-specific experts.}
\label{fig:fig4}
\end{figure}

Our framework is built upon Emotion-LLaMA~\cite{cheng2024emotion}, which takes as input a sequence of $m$ visual tokens $\mathcal{V}=\{\mathbf{v}_1, \dots, \mathbf{v}_m\}$, $n$ audio tokens $\mathcal{A} = \{\mathbf{a}_1, \dots, \mathbf{a}_n\}$, and $s$ fixed instruction text (prompt) tokens $\mathcal{T} = \{\mathbf{t}_1, \dots, \mathbf{t}_s\}$. 
To simplify notation, we denote the non-visual tokens as $\mathcal{N} = \{\mathcal{A}, \mathcal{T}\}$ and all tokens as $\mathcal{X} = \{\mathcal{V}, \mathcal{A}, \mathcal{T}\}$.
Given a video clip, the visual tokens $\mathcal{V}$ are extracted using three encoders: EVA-CLIP~\cite{sun2023eva} for global visual features, MAE~\cite{he2022masked} for local details, and VideoMAE~\cite{tong2022videomae} for temporal dynamics. The audio tokens $\mathcal{A}$ are encoded by HuBERT~\cite{hsu2021hubert}. The number of visual tokens is significantly larger than that of audio tokens ($m \gg n$), \eg, 256 \vs. 1 in Emotion-LLaMA. We find that this disparity in token counts leads to a notable bias toward the audio modality. In this section, we propose two modules to address this issue: inserting modality-specific experts for parameter-efficient fine-tuning, and applying attention reallocation during inference.

\input{tabs/ar_algorithm}

\subsection{Modality-Specific Experts}

\input{tabs/DA-MER}

To promote balanced learning across modalities, we propose modality-specific experts (\ourtrain): mixture of LoRA~\cite{hu2022lora} modules designed to enhance the emotion cues from each modality, combined with a regularized routing mechanism that dynamically adjusts their contributions. Specifically, we design three experts:
\begin{itemize}[leftmargin=1em] 
    \item \textbf{Visual Expert} $\mathcal{E}_\mathsf{v}(\cdot)$, which processes visual tokens $\mathcal{V}$ to enhance cues that are often underutilized by the base model.
    \item \textbf{Non-Visual Expert}  $\mathcal{E}_\mathsf{n}(\cdot)$, which handles audio tokens and text tokens ($\mathcal{N}=\{\mathcal{A}, \mathcal{T}\}$).
    \item \textbf{Omni Expert}  $\mathcal{E}_\mathsf{o}(\cdot)$, which processes all tokens ($\mathcal{X}=\{\mathcal{V},\mathcal{A},\mathcal{T}\}$).
\end{itemize}
For any token $\mathbf{x} \in \mathcal{X} \subset \mathbb{R}^d$, we assign it to the corresponding expert. To enable parameter-efficient training, each expert is implemented as an \textit{asymmetric soft mixture of LoRAs}. Specifically, each expert shares a rank-reduction matrix, and is equipped with $N$ rank-expansion matrices. 
Take the visual expert $\mathcal{E}_\mathsf{v}(\cdot)$ as an example, the output for a visual token $\mathbf{v}$ is computed as:
\begin{equation}\label{eq:mose_forward}
    \mathcal{E}_\mathsf{v}(\mathbf{v})=\sum_{i=1}^N \alpha_{\mathsf{v},i} (\mathbf{v})B_{\mathsf{v},i}A_\mathsf{v}\mathbf{v},\quad \text{where}\ \alpha_\mathsf{v}(\mathbf{v})=\text{softmax}(W_\mathsf{v}\mathbf{v}).
\end{equation}
Here, $A_\mathsf{v} \in \mathbb{R}^{r\times d} (r \ll d)$ is the shared rank-reduction matrix, $B_{\mathsf{v},i} \in \mathbb{R}^{d\times r}$ is the rank-expansion matrix, and $\alpha_\mathsf{v}(\mathbf{v}) \in \mathbb{R}^N$ computes the combining scores for each matrix $B_{\mathsf{v},i}$.    

To fuse the outputs of the three experts, we introduce a modality routing mechanism that dynamically adjusts their contributions on a sample-wise manner. For each sample, we first compute the mean representations of the visual and non-visual tokens, denoted as $\bar{\mathbf{v}}$ and $\bar{\mathbf{n}}$, respectively.
We then pass the representations through an importance network $f(\cdot)$, implemented as a lightweight MLP. The routing score $\lambda \in [0,1]$ for visual tokens is computed as 
\begin{equation}\label{eq:lambda}
    \lambda = \frac{1}{2} + \epsilon \cdot \text{tanh}(f(\bar{\mathbf{v}};\bar{\mathbf{n}})),
\end{equation}
where $\epsilon \in [0, 0.5]$ serves as a regularization to prevent the model over-relying on any single modality.  Sec.~\ref{sec:ablation} shows that both excessively small and large values of $\epsilon$ lead to suboptimal performance. With $1 - \lambda$ assigned as the weight for non-visual tokens,  any input token $\mathbf{x} \in \mathcal{X}$ in each MLLM layer is computed as:
\begin{equation}
    \mathbf{y}=\mathsf{FFN}(\mathbf{x})+\mathcal{E}_\mathsf{o}(\mathbf{x})+\lambda\mathds{1}_{[\mathbf{x}\in \mathcal{V}]} \mathcal{E}_\mathsf{v}(\mathbf{x}) + (1-\lambda)\mathds{1}_{[\mathbf{x}\in \mathcal{N}]} \mathcal{E}_\mathsf{n}(\mathbf{x}),
\end{equation}
where $\mathsf{FFN}(\cdot)$ is the frozen Transformer layer of MLLM, and $\mathds{1}_{[\cdot]}$ is the indicator function to assign $\mathbf{x}$ to the corresponding expert.

\subsection{Attention Reallocation}
As shown in Sec.~\ref{sec:audio_bias}, intermediate MLLM layers attend disproportionately to audio tokens. A straightforward approach, such as PAI~\cite{liu2024paying}, shifts entire audio attention to visual tokens in a static manner. Unfortunately, this approach induces a trade-off between audio and visual modalities—gains on video-aligned subsets degrade audio-aligned performance (see Sec.~\ref{sec:ablation} for details). In contrast, we first identify attention heads that over-rely on the audio modality on a per-sample basis, and then reallocate their attention toward visual tokens. Empirical results in Sec.~\ref{sec:ablation} confirm that this procedure does not impair the use of audio cues for reasoning.

\noindent\textbf{Identifying biased attention heads.}
Let $h \in [H]$ denote the index of an attention head. At each decoding step, the attention weight assigned to token $\mathbf{x}$ by head $h$ at layer $\ell \in [L]$ is denoted as $\omega_{h}(\mathbf{x})$.\footnote{For clarity, we omit the layer index $\ell$ hereafter.}
Let $S_{h}(\omega,\mathcal{X})=\sum_{\mathbf{x}\in \mathcal{X}}[ \omega_{h}(\mathbf{x})]$ denote the total attention weight assigned  by head $h$ to the token set $\mathcal{X}$.
We use two metrics to locate biased heads: \textbf{(1)} head-level attention ratio $c_h(\omega)$, which is defined as the  ratio of total attention to audio tokens over that to visual tokens for head $h$:
\begin{equation}\label{eq:head-ratio}
    c_h(\omega)=\frac{S_{h}(\omega,\mathcal{A})}{S_{h}(\omega,\mathcal{V})}.
\end{equation} 
\textbf{(2)} layer-level attention ratio $c(\omega)$ , defined analogously to $c_h(\omega)$ but aggregated over all heads in a layer: 
\begin{equation}\label{eq:layer-ratio}
    c(\omega)=\frac{\sum_{h\in [H]}[S_{h}(\omega,\mathcal{A})]}{\sum_{h\in [H]}[S_{h}(\omega,\mathcal{V})]}.
\end{equation}
We consider a layer $\ell$ biased if its layer-level ratio $c(\omega)$ exceeds a threshold $\tau$. In that case, a head $h$ is identified as biased if its head-level ratio $c_h(\omega)$  exceeds the layer-level ratio $c(\omega)$.  Formally, 
\begin{equation}\label{eq:bias_define}
    \mathcal{H}_\mathsf{bias}=\{h|c(\omega)>\tau\ \text{and}\ c_h(\omega) > c(\omega)\}
\end{equation}
This allows us to refine attention at a finer granularity, rather than modulating the entire attention layers and heads.

\input{tabs/EMER}

\input{tabs/MER_DFEW}

\input{tabs/human_evaluation}

\noindent\textbf{Reallocating attention weights.} Given a biased head $h \in \mathcal{H}_\mathsf{bias}$, we redistribute a portion of its audio attention to the visual modality. 
Let $\omega'_{h}(\mathbf{x})$ denote the redistributed attention weights for token $\mathbf{x}$. 
The redistributed weights are constrained to satisfy: 
\begin{align}
    c_h(\omega')&=c(\omega), \label{eq:target}\\
    S_{h}(\omega',\{\mathcal{A,V}\})&=S_{h}(\omega,\{\mathcal{A,V}\}), \label{eq:mass_total} \\
    \frac{\omega'_{h}(\mathbf{a})-\omega_{h}(\mathbf{a})}{S_{h}(\omega',\mathcal{A}) - S_{h}(\omega,\mathcal{A})} &=\frac{\omega_{h}(\mathbf{a})}{S_{h}(\omega,\mathcal{A})},\ \forall \mathbf{a}\in \mathcal{A}, \label{eq:audio_propotion} \\
     \frac{\omega'_{h}(\mathbf{v})-\omega_{h}(\mathbf{v})}{S_{h}(\omega',\mathcal{V}) - S_{h}(\omega,\mathcal{V})}
    &=\frac{\omega_{h}(\mathbf{v})}{S_{h}(\omega,\mathcal{V})},\ \forall \mathbf{v}\in \mathcal{V}. \label{eq:video_propotion}
\end{align}
Eq.~(\ref{eq:target}) enforces that the head-level attention ratio after redistribution matches the original layer-level ratio.
Eq.~(\ref{eq:mass_total}) ensures that the total attention assigned to audio and visual tokens remains unchanged.
Eqs.~(\ref{eq:audio_propotion}) and (\ref{eq:video_propotion}) guarantee that attention is redistributed proportionally among audio and visual tokens, \textit{preserving their original intra-modality distribution}. The closed-form solution for Eq.~(\ref{eq:target}-\ref{eq:video_propotion}) are:
\begin{align}
 \omega'_{h}(\mathbf{a})&=\omega_{h}(\mathbf{a})\cdot \biggl(1-\frac{\Delta_{h}}{S_{h}(\omega,\mathcal{A})}\biggr),\ \forall \mathbf{a} \in \mathcal{A}, \label{eq:audio_update} \\
    \omega'_{h}(\mathbf{v})&=\omega_{h}(\mathbf{v})\cdot \biggl(1+\frac{\Delta_{h}}{S_{h}(\omega,\mathcal{V})}\biggr),\ \forall \mathbf{v} \in \mathcal{V}, \label{eq:video_update} \\
    \text{where}\ \Delta_{h} &=\frac{S_{h}(\omega,\mathcal{A}) - c(\omega)\,S_{h}(\omega,\mathcal{V})}{1 + c(\omega)}.
\end{align}
This procedure is summarized in Algorithm~\ref{alg:audio_reallocation}, and is repeated for all Transformer layers $\ell \in [L]$.

%% file: tabs/ar_algorithm.tex
\begin{algorithm}[t]
        \caption{Pipeline of Attention Reallocation (\ourinfer)}\label{alg:audio_reallocation}
        \begin{algorithmic}[1]
        \State \textbf{Given}: original attention weights $\omega$, threshold $\tau$ 
        \State Compute layer-level ratio $c(\omega)$ via Eq.~(\ref{eq:layer-ratio})
        \If{$c(\omega)>\tau$}
            \For{$h\in [H]$}{}
                \State Compute head-level ratio $c_h(\omega)$ via Eq.~(\ref{eq:head-ratio})
                \If{$c_h(\omega)>c(\omega)$}
                    \State Update $\omega_h'$ via Eq.~(\ref{eq:audio_update}) and Eq.~(\ref{eq:video_update})
                \Else
                    \State Update $\omega'_h=\omega_h$
                \EndIf
            \EndFor
        \Else
            \State Update $\omega'=\omega$
        \EndIf
        \State \textbf{Return}: reallocated attention weights $\omega'$ 
        \end{algorithmic}
\end{algorithm}

%% file: tabs/DA-MER.tex
\begin{table*}[t]
    \centering
    \caption{Performance (\%) of emotion reasoning on~\ourdataset. ``Acc.'' and ``Rec.'' denote accuracy and recall, respectively.}
    \label{tab:CA-MER}
\begin{adjustbox}{width=0.95\linewidth}
        \begin{tabular}{l c| ccc| ccc| ccc| ccc}
            \toprule
            & & \multicolumn{3}{c|}{\textbf{Video-Aligned}} & \multicolumn{3}{c|}{\textbf{Audio-Aligned}} & \multicolumn{3}{c|}{\textbf{Consistent}} & \multicolumn{3}{c}{\textbf{Overall}} \\
            \textbf{Model} & \textbf{Modality} & \textbf{Acc.} & \textbf{Rec.} & \textbf{Avg.} & \textbf{Acc.} & \textbf{Rec.} & \textbf{Avg.} & \textbf{Acc.} & \textbf{Rec.} & \textbf{Avg.} & \textbf{Acc.} & \textbf{Rec.} & \textbf{Avg.} \\
            \midrule
            SALMONN~\cite{tang2023salmonn}         & A+T  & 29.37 & 22.56 & 25.97 & 53.35 & 34.19 & 43.77 & 47.72 & 38.53 & 43.13 & 43.48 & 31.76 & 37.62 \\
            mPLUG-Owl3~\cite{ye2024mplug}          & V+T  & 40.71 & 42.01 & 41.36 & 41.08 & 41.01 & 41.04 & 49.29 & 39.08 & 44.18 & 43.69 & 40.70 & 42.19 \\
            Chat-UniVi~\cite{jin2024chat}          & V+T  & 43.43 & 42.42 & 42.92 & 42.28 & 36.31 & 39.30 & 50.38 & 45.95 & 48.16 & 45.36 & 41.56 & 43.46 \\
            \midrule
            Vita1.5~\cite{fu2025vita}              & A+V+T& 49.36 & 46.60 & 47.98 & 57.51 & 48.15 & 52.83 & 57.17 & 52.52 & 54.84 & 54.68 & 49.09 & 51.88 \\
            Emotion-LLaMA~\cite{cheng2024emotion}  & A+V+T& 47.66 & 51.45 & 49.56 & 59.94 & 51.63 & 55.78 & 59.23 & 63.80 & 61.52 & 55.61 & 55.63 & 55.62 \\
            \rowcolor{Gray}
            \ourmethod                                & A+V+T& \textbf{58.42} & \textbf{54.28} & \textbf{56.35} & \textbf{65.33} & \textbf{55.26} & \textbf{60.30} & \textbf{67.68} & \textbf{65.87} & \textbf{66.77} & \textbf{63.81} & \textbf{58.47} & \textbf{61.14} \\
            \bottomrule
        \end{tabular}
    \end{adjustbox}
\end{table*}

%% file: tabs/EMER.tex
\begin{table}[t]
\centering
 \caption{Performance (\%) of emotion reasoning on EMER.}
\begin{adjustbox}{width=0.95\linewidth}
\tabstyle{10pt}
\begin{tabular}{lccc}
\toprule
\textbf{Model} & \textbf{Accuracy} & \textbf{Recall} & \textbf{Average} \\
    \midrule
    Otter ~\cite{li2023mimic}               & 34.43 & 24.39 & 29.41 \\
    Video-LLaVA ~\cite{lin2023video}           & 34.43 & 30.44 & 32.44 \\
    OneLLM ~\cite{han2024onellm}           & 38.19 & 30.01 & 34.10 \\
    PandaGPT ~\cite{su2023pandagpt}         & 36.59 & 32.89 & 34.74 \\
    Video-LLaMA ~\cite{zhang2023video}      & 38.59 & 35.54 & 37.06 \\
    Qwen-Audio ~\cite{chu2023qwen}          & 50.55 & 30.21 & 40.38 \\
    Video-ChatGPT ~\cite{maaz2023video}     & 46.13 & 35.05 & 40.59 \\
    VideoChat2 ~\cite{li2024mvbench}        & 51.98 & 35.99 & 43.98 \\
    SALMONN ~\cite{tang2023salmonn}         & 46.88 & 44.95 & 45.91 \\
    LLaMA-VID ~\cite{li2024llama}           & 47.59 & 44.63 & 46.11 \\
    Chat-UniVi ~\cite{jin2024chat}           & 49.78 & 44.37 & 47.08 \\
    VideoChat ~\cite{li2023videochat}       & 48.53 & 45.74 & 47.13 \\
    mPLUG-Owl ~\cite{ye2023mplug}           & 49.67 & 46.03 & 47.85 \\
    Emotion-LLaMA ~\cite{cheng2024emotion}   &  52.22 & 50.79 & 51.51 \\
    \rowcolor{Gray}
    \ourmethod   & \textbf{66.77} & \textbf{54.39} & \textbf{60.58} \\
\bottomrule
\end{tabular}
\end{adjustbox}
\label{tab:EMER}
\end{table}

%% file: tabs/MER_DFEW.tex
\begin{table*}[t!]
\centering
\setlength{\tabcolsep}{3.5pt}
\captionsetup[table]{labelsep=space,singlelinecheck=off}

\begin{minipage}[t]{0.31\textwidth}
\centering
\caption{F1 score of emotion recognition on MER2023}
\scalebox{0.95}[1]{  
\begin{tabular}{lc}
\toprule
\textbf{Method}  & \textbf{F1 Score} \\
\midrule
VideoMAE ~\cite{tong2022videomae}    & 0.6068 \\
HuBERT ~\cite{hsu2021hubert}         & 0.8511 \\
MER2023-Baseline ~\cite{lian2023mer} & 0.8675 \\
SSL-Transformer ~\cite{chen2023semi}       & {0.8853}\\
FBP ~\cite{cheng2023semi}            & 0.8855 \\
VAT ~\cite{ding2023learning}         & 0.8911 \\
Emotion-LLaMA ~\cite{cheng2024emotion} & 0.8087 \\
\rowcolor{Gray}
\ourmethod~                              & \textbf{0.9027} \\
\bottomrule
\end{tabular}%
}
\label{tab:MER2023}
\end{minipage}
\hfill
\begin{minipage}[t]{0.66\textwidth}
 \centering
\caption{Performance (\%) of emotion recognition on DFEW. ``UAR'' and ``WAR'' stands for  unweighted and weighted average recall, respectively. }
\scalebox{0.95}[1]{  
\begin{tabular}{lccccccc|cc}
\toprule
\multicolumn{1}{l}{\textbf{Method}} & \textbf{Hap.} & \textbf{Sad.} & \textbf{Neu.} & \textbf{Ang.} & \textbf{Sur.} & \textbf{Dis.} & \textbf{Fea.} & \textbf{UAR} & \textbf{WAR} \\
\midrule
EC-STFl ~\cite{jiang2020dfew}      & 79.18 & 49.05 & 57.85 & 60.98 & 46.15 & 2.76  & 21.51 & 45.35 & 56.51 \\
Former-DFER ~\cite{zhao2021former} & 84.05 & 62.57 & 67.52 & 70.03 & 56.43 & 3.45  & 31.78 & 53.69 & 65.70 \\
IAL ~\cite{li2023intensity}        & 87.95 & 67.21 & 70.10 & 76.06 & 62.22 & 0.00  & 26.44 & 55.71 & 69.24 \\
MAE-DFER ~\cite{sun2023mae}        & 92.92 & 77.46 & 74.56 & 76.94 & 60.99 & \textbf{18.62} & 42.35 & 63.41 & 74.43 \\
VideoMAE ~\cite{tong2022videomae}  & 93.09 & 78.78 & 71.75 & 78.74 & 63.44 & 17.93 & 41.46 & 63.60 & 74.60 \\
S2D ~\cite{chen2024static}         & 93.62 & \textbf{80.25} & \textbf{77.14} & \textbf{81.09} & 64.53 & 1.38  & 34.71 & 61.82 & \textbf{76.03} \\
Emotion-LLaMA ~\cite{cheng2024emotion} & 91.82 & 80.21 & 71.91 & 78.62 & 63.95 & 10.34 & 39.23 & 62.30 & 74.37 \\
\rowcolor{Gray}
\ourmethod~   & \textbf{93.87} & 74.41 & 73.97 & 80.00 & \textbf{67.35} & 13.79 & \textbf{46.41} & \textbf{64.26} & 75.61 \\
\bottomrule
\end{tabular}%
}
\label{tab:DFEW}
\end{minipage}
\vspace{-4mm}
\end{table*}

%% file: tabs/human_evaluation.tex
\begin{table}[h]
  \centering
  \caption{Human evaluation on \ourdataset.}
\begin{adjustbox}{width=1.0\linewidth}
  \begin{tabular}{lccc}
    \toprule
    \textbf{Method} & \textbf{V.-Aligned} & \textbf{A.-Aligned} & \textbf{Consist.} \\
    \midrule
        Emotion-LLaMA ~\cite{cheng2024emotion}   & 5.65 & 6.39 & 6.83 \\
        \rowcolor{Gray}
        \ourmethod                         & 6.00 & 7.16 & 7.61 \\
    \bottomrule
  \end{tabular}
  \label{tab:human_evaluation}
\end{adjustbox}
\vspace{-4mm}
\end{table}

%% file: section/6_expriments.tex
\section{Experiments}
\label{expriments}

\subsection{Setup}
\label{sec:dataset_setting}

\noindent\textbf{Tasks and datasets.}
We evaluate our \ourmethod~on both multimodal emotion reasoning and recognition tasks. \textbf{(1)} emotion reasoning requires the model to predict emotions with explanations. We adopt two datasets: EMER~\cite{lian2023explainable}, which contains 332 samples annotated with reasoning explanations, and our proposed \ourdataset. \textbf{(2)} emotion recognition, a single-label classification task, evaluated on MER2023~\cite{lian2023mer}, a multimodal emotion dataset featuring annotated TV/movie clips with visual, audio, and textual cues, and DFEW~\cite{jiang2020dfew}, a large-scale "in-the-wild" dynamic facial expression database consisting of over 16,000 video clips from thousands of movies.

\noindent\textbf{Evaluation metrics.} For emotion reasoning, following AffectGPT~\cite{lian2024affectgpt}, we use ChatGPT~\cite{ouyang2022training} to extract emotion-related keywords from the final conclusion of generated explanations. The keywords are clustered and compared with ground-truth to compute set-level accuracy and recall. For  MER2023~\cite{lian2023mer}, we report the F1 score, as recommended in prior work~\cite{lian2023mer,cheng2024emotion}. 
For DFEW~\cite{jiang2020dfew}, we measure Unweighted and Weighted Average Recall (UAR and WAR). 
See Appendix C.1 for the details of metrics. 

\input{tabs/ablation1}

\noindent\textbf{Implementation details.} 
We adopt the same base model, MiniGPT-v2~\cite{chen2023minigpt}, as used in Emotion-LLaMA~\cite{cheng2024emotion}. 
We also follow Emotion-LLaMA, to adopt the two-stage training strategy on the MERR dataset~\cite{cheng2024emotion}: pretraining on 28,618 coarse-quality data followed by fine-tuning on 4,487 high-quality data. 
Unlike Emotion-LLaMA, which trains separate models for emotion reasoning and recognition, our \ourmethod~optimizes a unified model for both tasks: at each training stage, tasks are interleaved at the batch level by randomly sampling either reasoning or recognition data.
The initial learning rate is set to $2\times 10^{-5}$ in the first stage and $1\times 10^{-5}$ in the second stage. Each stage is trained for 30 epochs, with 1000 iterations per epoch. A warm-up learning rate of $1\times 10^{-6}$ is applied, followed by cosine annealing for the subsequent epochs. For adaptation on the DFEW dataset, each epoch consists of 2000 iterations, and the learning rate is set to $5\times 10^{-5}$. We employ the AdamW~\cite{loshchilov2017decoupled} optimizer with a weight decay of $5\times 10^{-2}$. All experiments are conducted using four NVIDIA A800 GPUs.
For \ourtrain, we set $N=2$, the LoRA rank  $r=64$. For \ourinfer, the threshold is set to $\tau=1$.

\subsection{Comparison with State-of-the-Art Methods}
\label{sec:sota_comparison}


\noindent\textbf{Reasoning task.} Table~\ref{tab:CA-MER} presents the results on our \ourdataset~benchmark. We note several observations: \textbf{(1)} Incomplete-modality models (\ie, A+T or V+T) underperform on the subsets where missing modality conveys the true emotion in emotion conflicts. For example, SALMONN~\cite{tang2023salmonn} (A+T) excels on audio-aligned subsets but struggles with video-aligned ones, while Chat-UniVi~\cite{jin2024chat} (V+T) shows the opposite trend. \textbf{(2)} Models with complete modality inputs (A+V+T) achieve superior performance across all subsets compared to those with missing modalities. However, we observe a substantial performance drop on the video-aligned subset compared to the audio-aligned one, \eg, Emotion-LLaMA, the current SoTA, achieves 12\% lower accuracy on the video-aligned subset, indicating an audio bias in emotion conflicts. \textbf{(3)} Our \ourmethod~achieves the highest accuracy across all \ourdataset~subsets. Despite using the same training data and base model as Emotion-LLaMA, our \ourmethod~outperforms it by 6.79\%, 4.52\%, and 5.25\% in video-aligned, audio-aligned, and consistent scenarios, respectively. Furthermore, \ourmethod~reduces the performance gap between audio- and video-aligned subsets from 12\% to 6\%, demonstrating its bias-mitigation capability (see Table~\ref{tab:ablation_study2} for more evidence).
Table~\ref{tab:EMER} shows the reasoning performance on EMER~\cite{lian2023explainable}, where \ourmethod~achieves a SoTA score of 60.58\%. This highlights that \ourmethod~generalizes well beyond conflict scenarios.

\noindent\textbf{Recognition task.}
Table~\ref{tab:MER2023} summarizes the emotion recognition performance on MER2023~\cite{lian2023mer}. Our \ourmethod~ achieves the highest F1 score, surpassing the previous state-of-the-art Emotion-LLaMA~\cite{cheng2024emotion} by a remarkable 9.4\%. Table~\ref{tab:DFEW} reports the per-class accuracies (\ie, happy, sad, neutral, angry, surprise, disgust and fear), unweighted and weighted average recall on DFEW~\cite{jiang2020dfew}. Despite being designed for multi-task scenarios,  \ourmethod~still achieves the highest UAR (64.26\%), outperforming specialized single-task models.

\noindent\textbf{Human evaluation.}
We conducted a human study to assess the model’s consistency with human emotion understanding: for each \ourdataset~subset, 100 samples were randomly selected and rated (1–10 scale) by three annotators, blinded to model identity. As shown in the Table~\ref{tab:human_evaluation}, MoSEAR consistently receives higher scores than Emotion-LLaMA, indicating better human-perceived quality.

\subsection{Ablation Studies}
\label{sec:ablation}

\noindent\textbf{Study on the design of \ourtrain.} Three distinct designs of our \ourtrain~ are: {(1)} modality-specific modules — we design three experts for different token modalities; {(2)} \textit{asymmetric} soft mixture of LoRAs — each expert shares a rank-reduction matrix; and {(3)} regularized routing mechanism — a gating function that fuses cues from different modalities. To verify the effectiveness of the three designs, we compare our \ourtrain~(d) with several variants in Table~\ref{tab:ablation_study1}:  
\textbf{(a)} Modality-agnostic experts (\texttt{MoAE}): a standard mixture of LoRAs that takes all modalities as input, with comparable parameter size to our \ourtrain.
\textbf{(b)} Symmetric soft mixture of LoRAs: each expert contains multiple LoRAs with distinct rank-reduction matrices, leading to increased parameters.
\textbf{(c)} Modality fusion without routing: we replace the router with a simple average of the outputs from different experts.

Comparing Rows (a) and (d), we observe that modality-specific experts outperform the modality-agnostic variant, with gains of 2.52\% and 1.32\% on EMER and MER2023, respectively. Comparison between Rows (b) and (d) demonstrates that using a shared rank-reduction matrix yields better performance with fewer parameters. Row (c) highlights the importance of the gating mechanism, yielding an additional 1.09\% gain on EMER and 0.75\% on MER2023.
These findings justify the design of our three key modules.

\noindent\textbf{Effect of the hyper-parameter $\epsilon$.} The hyper-parameter $\epsilon \in [0, 0.5]$ in Eq.~(\ref{eq:lambda}) acts as a regularization term to prevent  over-reliance on single modality. We vary $\epsilon$ and report the performance on EMER and MER2023 in Appendix Figure 5. Note that $\epsilon=0$ corresponds to modality fusion with simple averaging and $\epsilon=0.5$ indicates routing without regularization. We find that both extreme choices of $\epsilon$ leads to suboptimal performance, while $\epsilon=0.1$, \ie~ $\lambda \in [0.4, 0.6]$, achieves the best trade-off.

\noindent\textbf{Study on our attention reallocation (\ourinfer).} To demonstrate the superiority of \ourinfer, we compare it with {PAI}~\cite{liu2024paying}, which mitigates bias by proportionally amplifying the attention weights assigned to visual tokens. We apply both attention modification methods to Emotion-LLaMA~\cite{cheng2024emotion} and our \ourtrain~models, and report the results in Table~\ref{tab:ablation_ar}. We observe a clear trade-off with PAI in emotion conflict scenarios: it improves performance on video-aligned samples but degrades it on audio-aligned ones, leading to stagnant or even lower scores on the consistent subset. 
We attribute the trade-off of PAI~\cite{liu2024paying} to two factors:
(i) it is coarse-grained, intervening at all heads and layers regardless of whether they exhibit bias. In contrast, our \ourinfer~targets only the heads with excessive audio bias (Eqs.~(\ref{eq:head-ratio}-\ref{eq:layer-ratio}));
(ii) it simply increases attention weights for visual tokens, which distorts the overall attention distribution. Instead, our \ourinfer~refines attention weights while preserving the original distribution structure (Eqs.~(\ref{eq:mass_total}-\ref{eq:video_propotion})).
In contrast, our \ourinfer~yields improvements across all datasets, with particularly large gains of 2.72\% and 2.42\% on the video-aligned subset when applied to Emotion-LLaMA and our \ourtrain, respectively.

\input{tabs/ablation2}

\input{tabs/threshold}

\noindent\textbf{Effect of the threshold $\tau$.}
The threshold $\tau$ in Eq.~(\ref{eq:bias_define}) determines whether a layer is biased. We vary $\tau = \{0,1,2,3\}$ and report the average accuracy and recall scores on \ourdataset~ in Table~\ref{tab:ablation_study_threhold}. Note that $\tau=0$ represents applying the adjustment to every layer without distinction, resulting in the worst performance. As $\tau$ increases, we observe that $\tau=1$ achieves the best performance.

\noindent\textbf{Effect of the number of experts ($N$).}
$N$ in Eq.~(\ref{eq:mose_forward}) controls the number of matrices $B$. We compare the performance with $N\in\{1, 2, 3\}$ and report the results on EMER and MER2023 in Appendix Table 15. We find that $N=2$ achieves the best performance, striking a balance between parameter efficiency and expressiveness. 

\noindent\textbf{Qualitative analysis.}
We conduct a separate qualitative analysis focusing on the role of \ourinfer~ and the outputs produced by our \ourmethod. (i) For \ourinfer~, we first compare the reasoning results in the video-aligned scenario, demonstrating that \ourinfer~ provides better reasoning outcomes compared to the counterpart. Next, in the audio-aligned scenario, we observe that PAI misleads attention and produces incorrect reasoning, whereas \ourinfer~ correctly infers the result. See Appendix D.2 for the visualization and more discussion. (ii) For \ourmethod, we compare its multimodal emotion reasoning outputs with Emotion-LLaMA on the video-aligned, audio-aligned, and consistent subsets, as well as on the EMER dataset. Our \ourmethod~ demonstrates strong reasoning abilities in hard cases. See Appendix D.3 for details.



%% file: tabs/ablation1.tex
\begin{table}[t]
  \centering
  \caption{Study on the design of \ourtrain.}
\begin{adjustbox}{width=0.95\linewidth}
  \begin{tabular}{ll ccc}
    \toprule
    & \textbf{Design} & \textbf{\#Param.} & \textbf{EMER} & \textbf{MER2023} \\
    \midrule
    (a) &\texttt{MoAE} &182M  & 56.84&88.95 \\
    (b)&\ourtrain~ w. symm. LoRAs &250M &57.44 &88.91 \\
    (c) &\ourtrain~ w.o. router &199M  & 58.27& 89.52\\
    \rowcolor{Gray}
    (d)&\ourtrain &200M & \textbf{59.36}&\textbf{90.27}  \\
    \bottomrule
  \end{tabular}
  \label{tab:ablation_study1}
\end{adjustbox}
\vspace{-4mm}
\end{table}

%% file: tabs/ablation2.tex
\begin{table}[t]
  \centering
  \caption{Study on the effect of \ourinfer. We report average accuracy and recall on \ourdataset.}\label{tab:ablation_ar}
  \begin{adjustbox}{width=1.\linewidth}
  \begin{tabular}{lccc|c}
    \toprule
    \multirow{2}{*}{\textbf{Design}}  & \multicolumn{3}{c|}{\textbf{\ourdataset}} & \multirow{2}{*}{\textbf{EMER}} \\
    &\textbf{V.-Aligned} & \textbf{A.-Aligned} & \textbf{Consist.} &  \\
    \midrule
    Emotion-LLaMA~\cite{cheng2024emotion}  &49.56  &55.78 &61.52 &51.51 \\
    \quad + PAI~\cite{liu2024paying} &52.08 ({\color{green}$\uparrow$}) &54.02 ({\color{red}$\downarrow$})&60.57 ({\color{red}$\downarrow$})&52.33 \\
    \quad + \ourinfer &52.28 ({\color{green}$\uparrow$}) &56.14 ({\color{green}$\uparrow$}) &62.11 ({\color{green}$\uparrow$}) &53.40 \\ \midrule
    \ourtrain &53.93 &59.55 &66.55 &59.36 \\
    \quad + PAI~\cite{liu2024paying} &56.82 ({\color{green}$\uparrow$}) &56.11 ({\color{red}$\downarrow$}) &66.11 ({\color{red}$\downarrow$}) &59.21 \\
    \quad + \ourinfer &56.35 ({\color{green}$\uparrow$}) &60.30 ({\color{green}$\uparrow$}) &66.77 ({\color{green}$\uparrow$}) &60.58 \\
    \bottomrule
  \end{tabular}
  \label{tab:ablation_study2}
  \end{adjustbox}
  \vspace{-4mm}
\end{table}

%% file: tabs/threshold.tex
\begin{table}[t]
  \centering
  \caption{Effect of the threshold $\tau$ of \ourinfer. We report the average of accuracy and recall on \ourdataset.}
  \tabstyle{6pt}
\begin{adjustbox}{width=0.95\linewidth}
  \begin{tabular}{ccccc}
    \toprule
    $\tau$ & \textbf{Video-Aligned} & \textbf{Audio-Aligned} & \textbf{Consistent} & \textbf{Overall}\\
    \midrule
    0 &55.32  & 58.54&66.72 &60.19 \\
    \rowcolor{Gray}
    1 &\textbf{56.35} &\textbf{60.30} &66.77 & \textbf{61.14}\\
    2 &56.14  & 59.58& \textbf{67.57} &61.09\\
    3 &55.97 & 59.21&67.26  &60.81  \\
    \bottomrule
  \end{tabular}
  \label{tab:ablation_study_threhold}
\end{adjustbox}
\vspace{-4mm}
\end{table}

%% file: section/7_conclusion.tex
\section{Conclusion}
\label{sec:conclusion}
In this paper, we present a systematic study of emotion MLLMs in the context of emotion conflicts. Our attention analysis on existing emotion MLLMs reveals a clear bias toward audio tokens, which impairs the integration of visual cues and results in inaccurate emotion reasoning. In addition, we find that the extreme imbalance between video and audio token counts is a key factor contributing to audio bias.
 To support evaluation in such scenarios, we introduce the Conflict-Aware Multimodal Emotion Reasoning (\ourdataset) dataset, consisting of three subsets targeting video-aligned, audio-aligned, and modality-consistent cases. To mitigate this bias, we propose \ourmethod, a novel framework comprising two key components: (1) Modality-specific experts (\ourtrain), which balance visual and non-visual modalities during training; and (2) Attention reallocation (\ourinfer), which calibrates the frozen model’s attention distribution during inference. Extensive experiments across multiple datasets and tasks demonstrate the effectiveness of \ourmethod~in mitigating audio bias and enhancing overall multimodal emotion reasoning.

%% file: section/X_appendix.tex
\clearpage
\appendix

\section{Closed-Form Solution of Attention Reallocation}
From Eq.~(\ref{eq:mass_total}), total attention weights on audio and visual tokens remain unchanged after reallocation. Thus, removing a mass of $\Delta_h$ from audio tokens requires adding the same amount to visual tokens:
\begin{equation}
S_{h}(\omega',\mathcal{A})
=
S_{h}(\omega,\mathcal{A}) - \Delta_{h}, 
\label{eq:mass_relation1}
\end{equation}
\begin{equation}
S_{h}(\omega',\mathcal{V})
=
S_{h}(\omega,\mathcal{V}) + \Delta_{h}.
\label{eq:mass_relation2}
\end{equation}
Plugging into Eqs.(\ref{eq:audio_propotion}) and (\ref{eq:video_propotion}) and rearranging terms, we obtain:
\begin{align}
\omega'_{h}(\mathbf{a})
&=
\omega_{h}(\mathbf{a}) 
\biggl(1 - \frac{\Delta_{h}}{S_{h}(\omega,\mathcal{A})}\biggr),
\quad
\forall \mathbf{a}\in\mathcal{A}. \label{eq:video_final}
 \\
\omega'_{h}(\mathbf{v})
&=
\omega_{h}(\mathbf{v}) 
\biggl(1 + \frac{\Delta_{h}}{S_{h}(\omega,\mathcal{V})}\biggr),
\quad
\forall \mathbf{v}\in\mathcal{V}.
\label{eq:audio_final}
\end{align}
From Eq.~(\ref{eq:target}), we have:
\begin{equation}
c_h(\omega') = \frac{
S_{h}(\omega',\mathcal{A})
}{
S_{h}(\omega',\mathcal{V})
}=c(\omega).
\label{eq:ratio_target}
\end{equation}
Combining with Eqs.~(\ref{eq:mass_relation1})~and~(\ref{eq:mass_relation2}), we have
\begin{equation}
\frac{
S_{h}(\omega,\mathcal{A}) - \Delta_{h}
}{
S_{h}(\omega,\mathcal{V}) + \Delta_{h}
}
=
c(\omega).
\end{equation}
Solving for \(\Delta_{h}\) gives
\begin{equation}
\Delta_{h}
=
\frac{
S_{h}(\omega,\mathcal{A}) - c(\omega)\,S_{h}(\omega,\mathcal{V})
}{
1 + c(\omega)
}.
\label{eq:delta}
\end{equation}

\section{\ourdataset~ Benchmark Details}
\label{sec:ca-mer_details}
 We introduce the prompts used during the construction of \ourdataset~ and visualization of subset construction process.
\subsection{Benchmark Construction Prompt}
\label{subsec:benchmark-prompt}
We begin by separately generating emotion reasoning for each modality (video and audio) using GPT-based models.

\noindent\textbf{Visual emotion reasoning generation.} For videos shorter than eight seconds, we sample at a rate of 1\,fps. For videos exceeding eight seconds, we uniformly sample eight frames. To help the language model capture detailed facial expressions, each sampled frame is enlarged to twice its original resolution in both width and height. We then employ a "gpt-4o" model with a carefully designed prompt to describe the emotions conveyed by the facial features and relevant visual context.

\begin{tcolorbox}[
    colback=gray!20, 
    colframe=black!70, 
    arc=2mm, 
    auto outer arc, 
    boxrule=0.3mm, 
    width=0.47\textwidth, 
    title=\centering Visual Emotion Reasoning Prompt
]
\textit{
You are an expert in human emotion analysis. 
First, analyze the main character. Then, focus on analyzing the person's facial expressions, including eyebrows, eyes, mouth, nose, facial muscles, etc. 
Finally, determine human emotions based on facial expressions.
The possible emotions include: neutral, happy, angry, worried, surprised, sad, fearful, doubtful, and contemptuous.
Please provide reasoning based on facial expressions in no more than 100 words.
}
\end{tcolorbox}

\noindent\textbf{Audio emotion reasoning generation.} We use a "gpt-4o-audio-preview" model to parse the corresponding audio segments. A specialized prompt guides the model to infer emotional attributes from acoustic characteristics such as intonation, rhythm, and volume.

\begin{tcolorbox}[
    colback=gray!20, 
    colframe=black!70, 
    arc=2mm, 
    auto outer arc, 
    boxrule=0.3mm, 
    width=0.47\textwidth, 
    title=\centering Audio Emotion Reasoning Prompt
]
\textit{
You are an expert in human emotion analysis. 
Describe the speaker's tone, speaking speed, and other vocal characteristics, and infer the speaker's emotions, which include: neutral, happy, angry, worried, surprised, sad, fear, doubt, or contempt.
Please provide your reasoning, and keep it under 100 words. Avoid uncertain answers like "I'm not sure" or "It’s difficult to judge." For example:\\
In the audio, the speaker speaks slowly, with a weak voice and a low pitch, indicating sadness.
}
\end{tcolorbox}

\noindent\textbf{Unimodal emotion label generation.} After obtaining unimodal emotion reasoning, we refine and consolidate the final emotion labels. We employ "gpt-4o" to analyze the descriptive cues from each modality, producing one explicit emotion category, including \textit{\{angry, happy, surprise, fear, sad, worry, neutral, doubt, contempt\}}. We use the following prompts to extract emotion labels:

\begin{tcolorbox}[
    colback=gray!20, 
    colframe=black!70, 
    arc=2mm, 
    auto outer arc, 
    boxrule=0.3mm, 
    width=0.47\textwidth, 
    title=\centering Emotion Label Generation Prompt
]
\textit{
Please summarize the character's final emotion based on the above emotion analysis. 
Choose one emotion from the following list: neutral, happy, angry, worried, surprised, sad, fearful, doubtful, or contemptuous. 
Please output only one word, without any additional content.\\
emotion analysis: \{analysis\}\\
emotion:
}
\end{tcolorbox}

\noindent\textbf{Majority voting.} We use a specific version of Emotion-LLaMA, which is exclusively trained on a larger-scale MER dataset for classification, to generate multimodal emotion labels. 
Then we divide the subsets and filter the data according to the majority voting method in Sec~\ref{sec:dataset}.

\noindent\textbf{Multimodal emotion reasoning generation.} 
We feed the visual and audio emotion cues, along with the emotion label, into gpt-4o to produce the final multimodal emotion reasoning process. Specifically, we use the following prompt to integrate the multimodal emotional cues. 

\subsection{Subset Samples Visualization}
\label{sec:subset_samples}
In this section, we select one representative sample from each of the three subsets to illustrate the characteristics of each category and the corresponding data construction process. 
Tables~\ref{tab:visual_data_template},~\ref{tab:audio_data_template},~\ref{tab:consistent_data_template} present the construction processes of the video-aligned, audio-aligned, and consistent samples, respectively. The Video Emotion Reasoning, Video Emotion Label, Audio Emotion Reasoning, Audio Emotion Label and Multimodal Emotion Reasoning are generated using the respective prompts introduced in the previous section~\ref{subsec:benchmark-prompt}. 

\begin{tcolorbox}[
    colback=gray!20, 
    colframe=black!70, 
    arc=2mm, 
    auto outer arc, 
    boxrule=0.3mm, 
    width=0.47\textwidth, 
    title=\centering Multimodal Emotion Reasoning Generation Prompt
]
\textit{
You are an expert in emotion analysis.  
We provide emotional clues from different modalities, including video, audio, and subtitles, along with a final emotional label. Your task is to analyze each clue (visual, audio, and textual), explain how they contribute to the inferred emotional state.\\
For each analysis, follow these steps:\\
1. Video Analysis: \{video analysis\}\\
2. Audio Analysis: \{audio analysis\}\\
3. Subtitle: \{subtitle\}\\
4. Carefully analyze the emotional tone from each clue. Your final explanation should align with the emotional label: \{final label\}.\\
Please ensure that your explanation covers the following:\\
1. Key emotional expressions in the visual, audio, and textual clues.\\
2. How these clues either support or conflict with each other.\\
3. A final, coherent emotional inference that aligns with the final label.\\
Answer in English. Your response should be concise and flow naturally, no more than 150 words.
}
\end{tcolorbox}

\input{appendix_tabs/video_aligned_data_sample}

\input{appendix_tabs/audio_aligned_data_sample}

\input{appendix_tabs/consistent_data_sample}

\section{Implementation Details}
\label{sec:implementation_details}

\subsection{Open Vocabulary Evaluation Metric}
\label{sec:evaluation_metric}
For our \ourdataset~  benchmark, we adopt the same evaluation metrics as EMER\cite{lian2023explainable}, using set-level accuracy and recall to assess the quality of open-vocabulary generation. Specifically, suppose that the ground-truth label set is $Y=\{y_i\}_{i=1}^{M}$
and the predicted label set is $\hat{Y}=\{\hat{y}_i\}_{i=1}^{N}$,where $M$ and $N$ denote the number of labels. Because the label space is not fixed, there may be synonyms among the labels (\ie, different expressions but the same meaning). Therefore, we first group all labels using "gpt-3.5-turbo-16k-0613" with the following prompt:
\begin{tcolorbox}[
    colback=gray!20, 
    colframe=black!70, 
    arc=2mm, 
    auto outer arc, 
    boxrule=0.3mm, 
    width=0.47\textwidth, 
    title=\centering OV-Emotion Label Group Prompt
]
\textit{
Please assume the role of an expert in the field of emotions. We provide a set of emotions. Please group the emotions, with each group containing emotions with the same meaning. Directly output the results. The output format should be a list containing multiple lists.
}
\end{tcolorbox}

Afterward, we employ the GPT-based grouping function  \(G(\cdot)\) to map each label to its corresponding group:
\begin{equation}
{Y}^{m} = \{G(x) |x \in \{{y}_i\}_{i=1}^M\}, \hat{Y}^{m} = \{G(x) |x \in \{\hat{y}_i\}_{i=1}^N\}.
\end{equation}

we then measure both set-level accuracy and recall, and subsequently average these two values to determine our final ranking metric:
\begin{equation}
\mbox{Accuracy} = \frac{|{Y}^{m} \cap \hat{{Y}}^{m}|}{|\hat{{Y}}^{m}|}, \mbox{Recall} = \frac{|{Y}^{m} \cap \hat{{Y}}^{m}|}{|{Y}^{m}|},
\end{equation}
\begin{equation}
\mbox{Average} = \frac{\mbox{Accuracy} + \mbox{Recall}}{2}.
\end{equation}

\subsection{Video-Audio Token Imbalance Training}
\label{sec:imbalance_training}
We present experiments on video-audio token imbalance. Specifically, we repeat the audio tokens along the sequence dimension to artificially increase their quantity, without introducing additional information. We conduct experiments by repeating the audio tokens 1, 50, 100, 200, and 256 times, respectively, such that the number of audio tokens becomes comparable to that of video tokens. As the number of audio tokens increases, the performance gap between the video-aligned and audio-aligned subsets consistently narrows. Notably, when the number of tokens from both modalities is equal, the model achieves better performance on the video-aligned subset.

\section{Experiments and Qualitative Analysis}
\label{sec:experiemnts_and_qualitative_analysis}

\subsection{Supplementary Experiments}
\label{sec:dfew_zeroshot_results}

\noindent\textbf{DFEW Zero-Shot Results.}In Table~\ref{tab:DFEW_zeroshot}, we present a comparison of the performance of zero-shot multimodal emotion recognition on the DFEW dataset.

\noindent\textbf{Study on our \ourinfer~ on hallucination mitigation.} We validate \ourinfer~ beyond emotion reasoning by applying it to hallucination mitigation for multimodal large language model(MLLM). As shown in the Table~\ref{tab:hallucination}, our method used on LLaVA-1.5~\cite{liu2024improved} achieves lower CHAIR scores on MSCOCO2014, demonstrating less hallucination compared to the leading de-hallucination baseline {PAI}~\cite{liu2024paying}.

\noindent\textbf{Running time.} 
We compare the FLOPs and inference time per sample of \ourmethod~ with Emotion-LLaMA~\cite{cheng2024emotion} on an NVIDIA A800 GPU. As shown in Table~\ref{tab:inference_time}, these results indicate that \ourmethod’s added complexity is minimal—only ~1\% higher FLOPs and negligible delay in wall-clock inference time—making it practical for real-world applications.

\subsection{Qualitative Analysis of Attention Reallocation}
\label{sec:qualitative_analysis_of_ar}
In this section, we qualitatively analyze the effects of \ourinfer~ on video-aligned and audio-aligned samples.

As shown in Table~\ref{tab:visual_sample1}, the speaker in the video adopts a neutral tone. Although the model without \ourinfer~ captures some cues related to a happy facial expression, the final prediction is influenced by the audio modality, leading the model to classify the emotional state as neutral. In contrast, the model with \ourinfer~ successfully mitigates the audio bias and correctly identifies the emotional state as happy.

As shown in Table~\ref{tab:audio_sample1}, the speaker in the video does not display any emotional expression. However, when the PAI method is applied in the intermediate layers of the model, it misguides the model into perceiving the person as anxious based on visual cues. On the other hand, when our \ourinfer~ method is employed, it does not mislead the model, which still classifies the emotional state as neutral.

\subsection{Multimodal Emotion Reasoning Comparison}
\label{sec:qualitative_analysis_of_merc}
This section presents a comparison of the performance of our \ourmethod~ model and the Emotion-LLaMA model on multimodal emotion reasoning tasks across four datasets. As shown in the Table~\ref{tab:visual_sample_comparison},~\ref{tab:audio_sample_comparison2},~\ref{tab:consistent_sample_comparison},~\ref{tab:EMER_sample_comparison}, the Emotion-LLaMA model often provides redundant yet incorrect reasoning processes, whereas our \ourmethod~ model outputs more concise and accurate explanations.

\input{tabs/DFEW_Zeroshot}

\input{tabs/hallucination}

\input{tabs/inference_time}

\input{tabs/experts_N}

\begin{figure}[!htbp]
\centering
\includegraphics[width=0.9\linewidth]{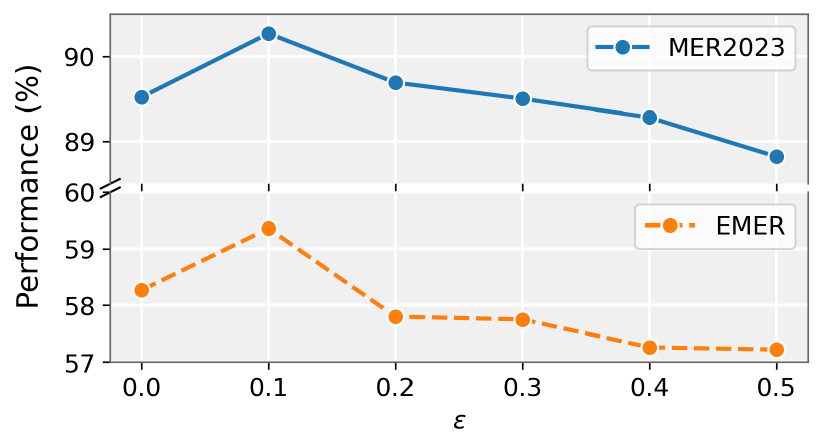}
\captionsetup{skip=6pt} 
\caption{Impact of the hyper-parameter $\epsilon$.}
\label{fig:fig5}
\end{figure}

\input{appendix_tabs/visual_ar1}

\input{appendix_tabs/audio_ar1}

\input{appendix_tabs/visual_aligned_sample1}

\input{appendix_tabs/audio_aligned_sample2}

\input{appendix_tabs/consistent_sample1}

\input{appendix_tabs/EMER_sample1}

%% file: appendix_tabs/video_aligned_data_sample.tex
\begin{table*}[!htbp]\centering
\caption{An Example of Video-Aligned Sample Generation}
\begin{minipage}{1.65\columnwidth}    \centering
\begin{tcolorbox} 
    \centering
      \footnotesize
    \begin{tabular}{p{0.98\columnwidth} c}
   { {} } & \hspace{-5cm} \multirow{5}{*}{ \includegraphics[height=2cm]{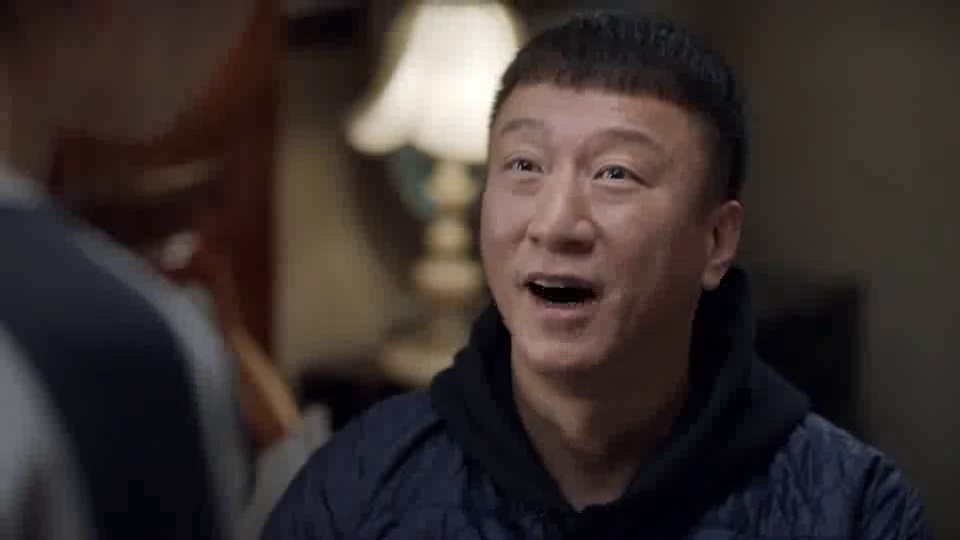} } \\
{ {\bf Audio Cues:} } & \\
rising intonation, disbelief, questioning tone  & \\
{ {\bf Visual Cues: } } & \\
slightly raised eyebrows, wide-open mouth & \\
{ {\bf Emotion Label: } } & \\
Happy\\
\hrulefill & \\
{ {\bf Visual Emotion Reasoning:} } & \\
In the provided video frames, the main character shows distinct shifts in emotion. Initially, the wide eyes and slightly raised eyebrows in the first frame suggest worry or surprise. In the second frame, the neutral facial expression suggests contemplation. Finally, the wide-open mouth and upward curved lips, along with the visible cheek raising in the third frame, indicate happiness. This progression reveals the character transitioning from worry or surprise to neutrality, and finally to happiness or relief.& \\
{ {\bf Visual Emotion Label: } } & \\
Happy\\
\hrulefill & \\
{ {\bf Audio Emotion Reasoning:} } & \\
In the audio, the speaker uses a rising intonation, which could indicate a challenge or disbelief, and a questioning tone that might suggest doubt or suspicion. The speaking speed seems moderate, adding a layer of sincerity or insistence to their speech. Based on these vocal cues, the emotion conveyed is doubt or skepticism.& \\
{ {\bf Audio Emotion Label: } } & \\
Doubtful\\
\hrulefill & \\
{ {\bf Multimodal Emotion Reasoning} } & \\
The visual clue shows a clear progression from worry/surprise to neutrality and eventually to happiness or relief, indicating a positive resolution of initial uncertainty. The audio, however, conveys doubt or skepticism through rising intonation and questioning tone, which introduces a more conflicted emotional undertone. The subtitle, "It's not working, is it," aligns with the skeptical audio tone, suggesting disappointment or uncertainty. To reconcile this, the transition to happiness seen visually suggests the character's emotional journey concludes positively, perhaps due to a resolution or realization that overrides earlier doubt. The contrast between the audio and visual clues implies the initial skepticism expressed vocally gives way to emotional relief or joy captured clearly in the final facial expression. The journey from doubt to happiness reflects that while the process starts with uncertainty, the ultimate emotional state is one of satisfaction or relief. Thus, happiness is the dominant emotion.& \\
\end{tabular}
\end{tcolorbox}
\vspace{-2mm}
\label{tab:visual_data_template}
\end{minipage}
\end{table*}

%% file: appendix_tabs/audio_aligned_data_sample.tex
\begin{table*}[!htbp]\centering
\caption{An Example of Audio-Aligned Sample Generation}
\begin{minipage}{1.65\columnwidth}    \centering
\begin{tcolorbox} 
    \centering
      \footnotesize
    \begin{tabular}{p{0.95\columnwidth} c}
   { {} } & \hspace{-5cm} \multirow{5}{*}{ \includegraphics[height=2cm]{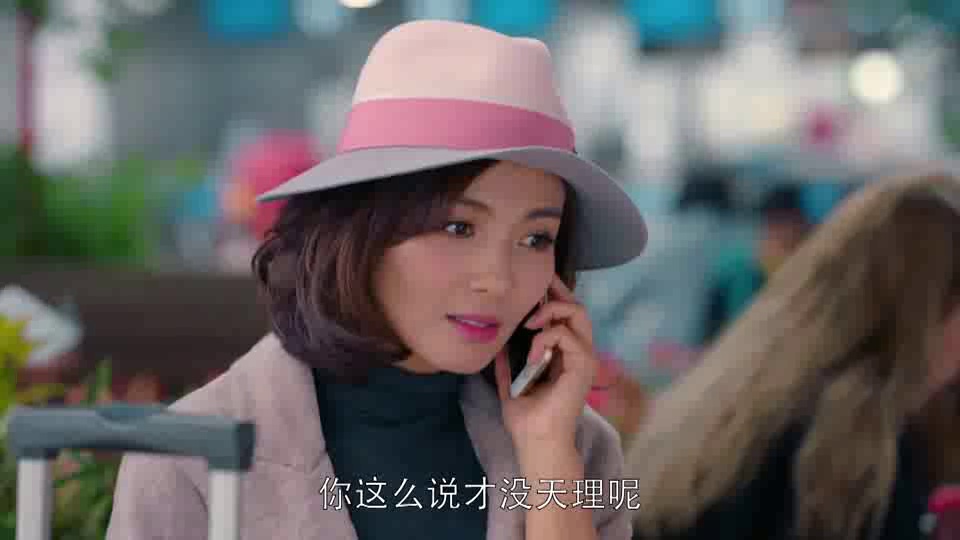} } \\
{ {\bf Audio Cues:} } & \\
assertive tone, moderate speaking speed & \\
{ {\bf Visual Cues: } } & \\
slight upward curve of the mouth, widened eyes, raised eyebrows  & \\
{ {\bf Emotion Label: } } & \\
Angry\\
\hrulefill & \\
{ {\bf Visual Emotion Reasoning:} } & \\
The main character exhibits a mix of emotions, primarily happy and slightly surprised. Her facial expressions display these emotions through a slight upward curve of the mouth, indicating happiness, and widened eyes and raised eyebrows, suggesting surprise. The gentle arch of her eyebrows and the relaxed facial muscles further confirm her mood. Overall, the visible expressions suggest that the character is engaged in an enjoyable or unexpected conversation, reflecting a generally positive and receptive emotional state.& \\
{ {\bf Visual Emotion Label: } } & \\
Happy\\
\hrulefill & \\
{ {\bf Audio Emotion Reasoning:} } & \\
In the audio, the speaker's tone is assertive and slightly confrontational, with a moderate speaking speed. The voice has a firm and clear quality, indicating a sense of dissatisfaction and frustration. The speaker's choice of words and the way they are delivered suggest that the emotion conveyed is anger.& \\
{ {\bf Audio Emotion Label: } } & \\
Angry\\
\hrulefill & \\
{ {\bf Multimodal Emotion Reasoning} } & \\
The visual clue suggests happiness and slight surprise, conveyed through smiling and widened eyes, reflecting a generally positive and receptive state. This contrasts sharply with the audio clue, where the speaker’s firm tone, moderate pace, and assertive delivery convey dissatisfaction and frustration, commonly associated with anger. The subtitle reinforces this, as the words express sarcasm and exasperation, indicating irritation or annoyance. The textual content aligns closely with the audio but conflicts with the cheerful and surprised expressions seen in the video. A coherent emotional interpretation integrates these aspects. The visual happiness and surprise may reflect an initial reaction or an external mask, while the tone and words reveal the speaker’s true emotional state: anger. The inconsistency likely stems from a deliberate attempt to disguise annoyance with outward positivity, a dynamic often seen in social interactions. Despite the mixed signals, the underlying anger emerges as the dominant emotion. & \\
\end{tabular}
\end{tcolorbox}
\vspace{-2mm}
\label{tab:audio_data_template}
\end{minipage}
\end{table*}

%% file: appendix_tabs/consistent_data_sample.tex
\begin{table*}[!htbp]\centering
\caption{An Example of Consistent Sample Generation}
\begin{minipage}{1.65\columnwidth}    \centering
\begin{tcolorbox} 
    \centering
      \footnotesize
    \begin{tabular}{p{0.95\columnwidth} c}
   { {} } & \hspace{-5cm} \multirow{5}{*}{ \includegraphics[height=2cm]{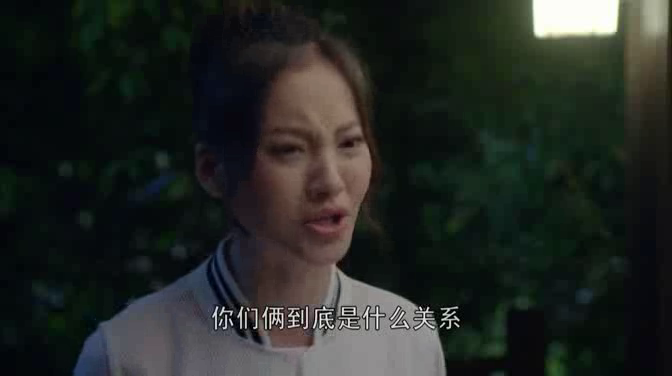} } \\
{ {\bf Audio Cues:} } & \\
raised voice, intensity and urgency speech & \\
{ {\bf Visual Cues: } } & \\
tightened mouth, clenched jaw, furrowed eyebrows  & \\
{ {\bf Emotion Label: } } & \\
Angry\\
\hrulefill & \\
{ {\bf Visual Emotion Reasoning:} } & \\
The main character exhibits emotions of anger and frustration. The analysis of the person's facial expressions reveals a tightened area around the mouth and a clenched jaw, indicative of anger. The eyebrows may be furrowed, further expressing displeasure. The overall facial tension and the context provided by the subtitles convey a scenario where the character is demanding honesty and clarification, leading to the conclusion that the person is likely feeling angry and frustrated.& \\
{ {\bf Visual Emotion Label: } } & \\
Angry\\
\hrulefill & \\
{ {\bf Audio Emotion Reasoning:} } & \\
In the audio, the speaker's tone is demanding and forceful, with a raised voice that suggests frustration or anger. The intensity and urgency in their speech, coupled with a fast speaking speed, point towards an emotional state of anger or frustration, likely due to seeking answers or clarity on a matter they find important.& \\
{ {\bf Audio Emotion Label: } } & \\
Angry\\
\hrulefill & \\
{ {\bf Multimodal Emotion Reasoning} } & \\
The visual clue reveals clear signs of anger and frustration through tightened lips, a clenched jaw, and furrowed eyebrows, all of which reflect tension and displeasure. These expressions align with someone experiencing strong negative emotions, potentially due to dissatisfaction or confrontation. The audio clue further supports this emotional state; the demanding, raised tone of voice, combined with forceful and urgent speech, strongly conveys frustration. The speaker's vocal delivery suggests an emotional investment in obtaining clarity or answers, consistent with the visual expressions. The text, "Can you honestly tell me today who you two are," adds context to the emotional tone. The directness and urgency of the question highlight frustration and a desire for truth, complementing the anger expressed visually and audibly. Together, the clues harmonize, converging on the emotional inference of anger driven by unmet expectations or a need for honesty. All modalities strongly reinforce this conclusion. & \\
\end{tabular}
\end{tcolorbox}
\vspace{4mm}
\label{tab:consistent_data_template}
\end{minipage}
\end{table*}

%% file: tabs/DFEW_Zeroshot.tex
\begin{table*}[htbp]
\centering
\renewcommand\tabcolsep{1pt}
\renewcommand\arraystretch{1.06}
\caption{Zero-shot multimodal emotion recognition on DFEW.}
\scalebox{0.9}{
\begin{tabular}{lccccccc|ccc}
\toprule
{Method}  & \textbf{Hap} & \textbf{Sad} & \textbf{Neu} & \textbf{Ang} & \textbf{Sur} & \textbf{Dis} & \textbf{Fea} & \textbf{UAR} & \textbf{WAR}\\
\midrule
Qwen-Audio ~\cite{chu2023qwen}          & {25.97}  & {12.93}  & {67.04}  & {29.20}  & {6.12}  & {0.00}  & {35.36}  & {25.23}  & {31.74}\\
LLaVA-NEXT ~\cite{liu2024llava}          & {57.46}  & \textbf{79.42}  & {38.95}  & {0.00}   & {0.00}  & {0.00}  & {0.00}   & {25.12}  & {33.75}\\
MiniGPT-v2 ~\cite{chen2023minigpt}          & \textbf{84.25}  & {47.23}  & {22.28}  & {20.69}  & {2.04}  & {0.00}  & {0.55}   & {25.29}  & {34.47}\\
Video-LLaVA(image) ~\cite{lin2023video}  & {37.09}  & {27.18}  & {26.97}  & {58.85}  & {12.97} & {0.00}  & {3.31}   & {20.78}  & {31.10}\\
Video-LLaVA(video) ~\cite{lin2023video}  & {51.94}  & {39.84}  & {29.78}  & {58.85}  & {0.00}  & {0.00}  & {2.76}   & {26.17}  & {35.24}\\
Video-LlaMA ~\cite{zhang2023video}         & {20.25}  & {67.55}  & \textbf{80.15}  & {5.29}   & {4.76}  & {0.00}  & {9.39}   & {26.77}  & {35.75}\\
GPT-4V ~\cite{lian2024gpt}              & {62.35}  & {70.45}  & {56.18}  & {50.69}  &{32.19} & \textbf{10.34} & \textbf{51.11}  & \textbf{47.69}      & {54.85}\\

Emotion-LLaMA ~\cite{cheng2024emotion} & {70.76}  & {79.68}  & {32.96}  & {39.08}  & {41.84} & {0.00}  & {0.00}  & {37.76}  & {47.71}\\

\rowcolor{Gray}
\ourmethod & {79.35}  & {75.20}  & {40.45}  & \textbf{69.66}  & \textbf{42.86} & {0.00}  & {3.87}  & {44.48}  & \textbf{56.60}\\
\bottomrule
\end{tabular}
}
\label{tab:DFEW_zeroshot}
\end{table*}

%% file: tabs/hallucination.tex
\begin{table}[!htbp]
  \centering
  \caption{Study on the effect of \ourinfer~ on hallucination mitigation task. We report $\textbf{CHAIR}_{\textbf{s}}$ and $\textbf{CHAIR}_{\textbf{i}}$ on MSCOCO2014.}
  \tabstyle{15pt}
\begin{adjustbox}{width=0.95\linewidth}
  \begin{tabular}{lcc}
    \toprule
    \textbf{Method} & $\textbf{CHAIR}_{\textbf{s}}$ $\downarrow$ & $\textbf{CHAIR}_{\textbf{i}}$ $\downarrow$ \\
    \midrule
    LLaVA-1.5~\cite{liu2024improved}        & 0.498 & 0.140 \\
    \quad + PAI~\cite{liu2024paying}        & 0.328 & 0.090 \\
    \rowcolor{Gray}
    \quad + \ourinfer~       & \textbf{0.286} & \textbf{0.085} \\
    \bottomrule
  \end{tabular}
  \label{tab:hallucination}
\end{adjustbox}
\end{table}

%% file: tabs/inference_time.tex
\begin{table}[!htbp]
  \centering
  \caption{Running time. We report FLOPs and inference time per sample on \ourdataset.}
\begin{adjustbox}{width=0.95\linewidth}
  \begin{tabular}{lccc}
    \toprule
    \textbf{Model} & \textbf{FLOPs (T)} & \textbf{Inference Time (ms)} \\
    \midrule
    Emotion-LLaMA~\cite{cheng2024emotion} & 4.60 & 14.99 \\
    \ourtrain~        & 4.65 & 17.86 \\
    \ourmethod~        & 4.65 & 19.90 \\
    \bottomrule
  \end{tabular}
  \label{tab:inference_time}
\end{adjustbox}
\end{table}

%% file: tabs/experts_N.tex
\begin{table}[!htbp]
  \centering
  \caption{Effect of the number of experts $N$. }
  \tabstyle{15pt}
\begin{adjustbox}{width=0.9\linewidth}
  \begin{tabular}{cccc}
    \toprule
    {$N$} & \textbf{\#Param.} & \textbf{EMER} & \textbf{MER2023} \\
    \midrule
    1 &148M  & 56.32&88.66 \\
    \rowcolor{Gray}
    2 &200M &\textbf{59.36} &\textbf{90.27} \\
    3 &250M  & 58.85& 89.52\\
    \bottomrule
  \end{tabular}
  \label{tab:ablation_study_N_experts}
\end{adjustbox}
\end{table}

%% file: appendix_tabs/visual_ar1.tex
\begin{table*}[ht]
\begin{minipage}{0.98\textwidth}
\caption{A video-aligned example of multimodal emotion reasoning comparing \ourtrain~ with \ourmethod.}
\centering  
\scalebox{0.9}{
    \begin{tabular}{l|p{16cm} }
\toprule
\multicolumn{2}{l}{\bf Video-Aligned Example of Multimodal Emotion Reasoning} \\
\midrule
& {\includegraphics[height=3cm]{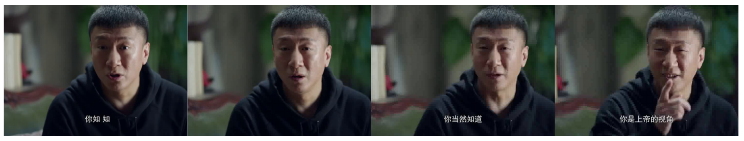}} \\
Subtitle (En) & {Oh, of course you know, you are God’s perspective.} \\
\midrule
Ground Truth & {The video clue shows a progression from concern (raised eyebrows, furrowed brow) to happiness (slight smile, softened eyes), suggesting an emotional shift. The initial worry transitions to relief or joy. The audio clue conveys a normal, slightly playful tone with possible sarcasm, pointing to mixed emotions like neutral or superiority. The textual clue, "Oh, of course you know, you are God’s perspective," implies sarcasm, adding a layer of mockery or playful disbelief. These clues initially appear conflicting, with visual worry, audio playfulness, and textual sarcasm. However, they converge if interpreted as a moment where the speaker, though voicing sarcasm, transitions emotionally to lighthearted amusement. The sarcastic comment might stem from an underlying sense of ironic humor, aligning with the shift in facial expression to happiness. By integrating all modalities, the overall emotional conclusion is happiness, refined by a mix of humor and relief.}  \\
\midrule
\ourtrain & {In the video, the individual's facial expression is characterized by a cheek raiser, indicating amusement or playfulness. Their tone and intonation are normal, and their statement is matter-of-fact, suggesting a neutral emotional state.}  \\
\midrule
\ourmethod & {In the video, the individual's facial expression is characterized by a cheek raiser, indicating amusement or playfulness. Their tone and intonation are normal, and they speak in a casual and relaxed manner. The combination of their facial expression and tone suggests that they are feeling amused or playful}.  \\
\midrule
\end{tabular}
}
\vspace{5mm}
\label{tab:visual_sample1}
\end{minipage}
\end{table*}

%% file: appendix_tabs/audio_ar1.tex
\begin{table*}[ht]
\begin{minipage}{0.98\textwidth}
\caption{An audio-aligned example of multimodal emotion reasoning comparing \ourmethod~ with \ourtrain+PAI.}
\centering  
\scalebox{0.90}{
    \begin{tabular}{l|p{16cm} }
\toprule
\multicolumn{2}{l}{\bf Audio-Aligned Example of Multimodal Emotion Reasoning} \\
\midrule
& {\includegraphics[height=3cm]{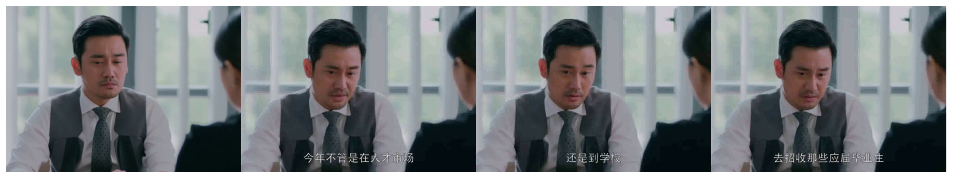}} \\
Subtitle (En) & {Well, this year...we will recruit those who welcome graduates.} \\
\midrule
Ground Truth & {The visual clue suggests the main character is experiencing concern or worry, indicated by furrowed eyebrows, a downward mouth tilt, and narrowed eyes. These features typically convey feelings of doubt or focus. In contrast, the audio presents a calm and steady tone, suggesting neutrality. The speaker's moderate speed and clear voice indicate a composed discussion without strong emotional undertones. The subtitle, "Well, this year...we will recruit those who welcome graduates," appears factual and lacks emotive language, aligning with a neutral stance. Although the visual cue of worry might initially seem at odds with the neutral tone conveyed by both audio and subtitle, it is plausible that the character is masking worry with a composed exterior. Thus, given the strength of the audio and textual neutrality, the overall emotional state inferred is neutral, possibly indicating the character's attempt to maintain composure despite underlying concerns.}  \\
\midrule
\ourtrain+PAI & {In the video, a man sits at a conference table with a facial expression characterized by a chin raiser and a lip tightener, indicating a sense of tension or unease. His tone and intonation are normal, but his facial expression suggests he may be feeling anxious or uncomfortable.}  \\
\midrule
\ourmethod & {In the video, a man sits at a desk with a facial expression characterized by a brow lowered and lid tightener, indicating focused attention. His tone and intonation are normal, and his speech is straightforward and matter-of-fact, suggesting a neutral emotional state.}  \\
\midrule
\end{tabular}
}
\vspace{5mm}
\label{tab:audio_sample1}
\end{minipage}
\end{table*}

%% file: appendix_tabs/visual_aligned_sample1.tex
\begin{table*}[ht]
\begin{minipage}{0.98\textwidth}
\caption{A video-aligned example of multimodal emotion reasoning comparing Emotion-LLaMA with \ourmethod.}
\centering  
\scalebox{0.90}{
    \begin{tabular}{l|p{16cm} }
\toprule
\multicolumn{2}{l}{\bf Video-Aligned Example of Multimodal Emotion Reasoning} \\
\midrule
& {\includegraphics[height=3cm]{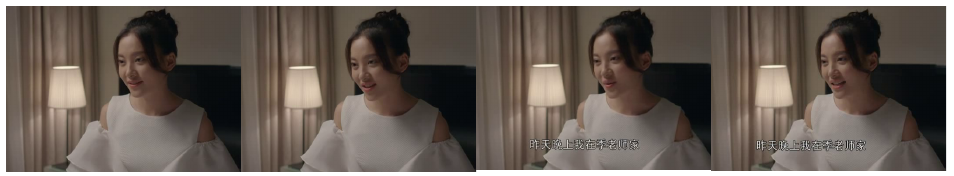}} \\
Subtitle (En) & {Last night, I was at Teacher Qi's house.} \\
\midrule
Ground Truth & {The visual clue strongly suggests happiness, as evidenced by the character’s relaxed facial muscles, squinted eyes, and a genuine smile. This facial expression aligns with a joyous emotional state. However, the audio clue reflects neutrality, with a calm, steady tone and moderate speed, indicating composure rather than overt excitement or joy. The textual clue refers to a past event at "Teacher Qi's house," which doesn't explicitly convey an emotional tone, leaving it open to interpretation based on visual and auditory cues. The visual indication of happiness outweighs the neutral tone in the audio. It’s plausible the character is reflecting on a pleasant memory, internally happy despite not vocalizing the emotion. Emotional subtlety, where happiness isn't audibly expressed but visually apparent, resolves the inconsistency. Thus, integrating all clues, it is logical to conclude that the dominant emotional state is happiness, displayed more prominently through facial expressions than verbal or vocal cues.}  \\
\midrule
Emotion-LLaMA & {In the video, a woman sits in a living room, wearing a white dress, and her facial expression is neutral, with her eyes looking downward her tone and intonation are normal, and she speaks in a matter-of-fact manner, stating, last night, i was at teacher qi's house the context of the conversation suggests that she is simply sharing information about her recent social activity, without any apparent emotional response the lack of emotional cues and the neutral tone of her voice indicate that the woman is likely feeling a sense of calmness or indifference, rather than any strong emotions.}  \\
\midrule
\ourmethod & {In the video, a woman sits on a couch, wearing a white dress, with a big smile and dimpled expression, indicating a strong sense of joy. Her tone and intonation are cheerful, and she says, "Last night, I was at Teacher Qi's house." This combination of facial expressions, tone, and context suggests that the person is feeling happy.}  \\
\midrule
\end{tabular}
}
\vspace{5mm}
\label{tab:visual_sample_comparison}
\end{minipage}
\end{table*}

%% file: appendix_tabs/audio_aligned_sample2.tex
\begin{table*}[ht]
\begin{minipage}{0.98\textwidth}
\caption{An audio-aligned example of multimodal emotion reasoning comparing Emotion-LLaMA with \ourmethod.}
\centering  
\scalebox{0.90}{
    \begin{tabular}{l|p{16cm} }
\toprule
\multicolumn{2}{l}{\bf Audio-Aligned Example of Multimodal Emotion Reasoning} \\
\midrule
& {\includegraphics[height=3cm]{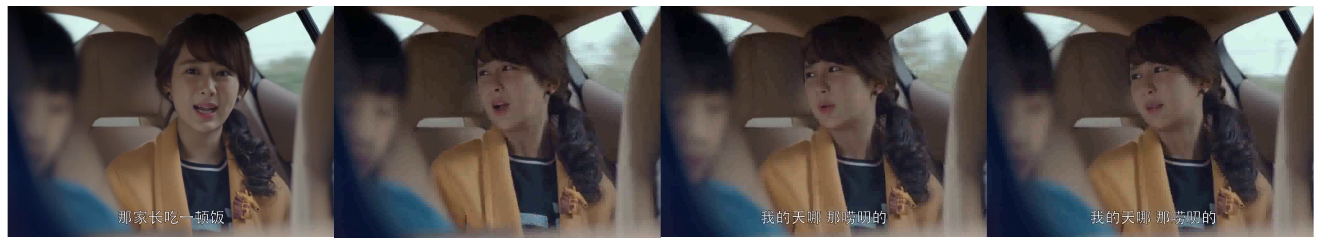}} \\
Subtitle (En) & {Parent, Have a meal, that day.} \\
\midrule
Ground Truth & {In the video, the main character's furrowed brows, wide eyes, and tense mouth convey doub. Her sideways glance accentuates hesitation and contemplation, suggesting she is unsettled by her circumstances. The audio reveals a tonal mix of surprise and concern, particularly through the phrase "aiya," which implies mild exasperation. Despite its light and casual tone, there is an undercurrent of anxiety. The subtitle, "Parent, Have a meal, that day," is neutral but implies a context that may contribute to the character's unease, possibly related to an expected event or interaction. The audio clues indicate a mix of surprise and worry, aligning with the inferred emotional state of being worried. While the subtitle itself does not directly convey emotion, it suggests a scenario that could underlie the character's anxiety, thus harmonizing with the worried emotion.}  \\
\midrule
Emotion-LLaMA & {In the video, a woman sits in a car, smiling and looking directly at the camera, displaying a happy and content expression her tone and intonation are normal, and she speaks in a normal tone, saying, parent, have a meal, that day the context of the conversation suggests a casual, everyday exchange, with no apparent emotional distress or conflict the woman's facial expression and tone of voice, combined with her normal speech patterns, indicate a positive and relaxed emotional state, likely reflecting her happiness and contentment.}  \\
\midrule
\ourmethod & {In the video, the person's facial expressions, including lowered eyebrows, widened eyes, and lifted chin, suggest a strong emotional response. The tone and intonation of their voice, particularly the fearful quality, further reinforces this inference. The phrase "Have a meal, that day" implies a sense of urgency or concern, which may be causing the person's emotional distress.}  \\
\midrule
\end{tabular}
}
\vspace{5mm}
\label{tab:audio_sample_comparison2}
\end{minipage}
\end{table*}

%% file: appendix_tabs/consistent_sample1.tex
\begin{table*}[ht]
\begin{minipage}{0.98\textwidth}
\caption{A consistent example of multimodal emotion reasoning comparing Emotion-LLaMA with \ourmethod.}
\centering  
\scalebox{0.90}{
    \begin{tabular}{l|p{16cm} }
\toprule
\multicolumn{2}{l}{\bf Consistent Example of Multimodal Emotion Reasoning} \\
\midrule
& {\includegraphics[height=3cm]{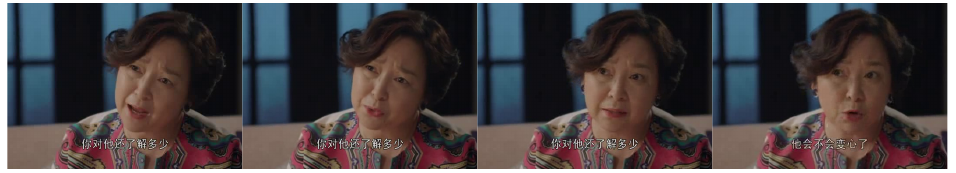}} \\
Subtitle (En) & {How much do you still know about him?} \\
\midrule
Ground Truth & {The facial expressions in the video suggest a worried state, as seen in the furrowed eyebrows and slightly squinted eyes, which signal concern or determination. The open mouth and tense upper lip indicate focus or urgency, reinforcing a sense of serious engagement with an underlying issue. The audio clue, characterized by a steady yet slightly tense tone and a hint of urgency, echoes this concern, suggesting an anxious need to gather or clarify information. The subtitle, "How much do you still know about him?" conveys a direct and pressing question, further emphasizing the speaker's worry about the level of understanding or potential consequences of the knowledge being referenced. All three modalities—visual, auditory, and textual—work cohesively to reveal an emotional state of worry, stemming from a pressing need for clarity and the potential emotional impact tied to the answer.}  \\
\midrule
Emotion-LLaMA & {In the video, a woman sits on a couch, looking directly at the camera, and her facial expression is a mixture of surprise and curiosity her tone and intonation are normal, and she asks, how much do you still know about him? the context suggests that she is discussing someone she is familiar with, and her question implies a sense of curiosity about their current situation or circumstances the combination of her facial expression and the tone of her voice indicates that she is intrigued and wants to know more about the person, which is a sign of a neutral emotion.}  \\
\midrule
\ourmethod & {In the video, a woman sits on a couch, looking directly at the camera with a furrowed brow, indicating a sense of concern or unease. Her facial expression and tone suggest she is asking a question or seeking clarification, implying she is worried about something.}  \\
\midrule
\end{tabular}
}
\vspace{5mm}
\label{tab:consistent_sample_comparison}
\end{minipage}
\end{table*}

%% file: appendix_tabs/EMER_sample1.tex
\begin{table*}[ht]
\begin{minipage}{0.98\textwidth}
\caption{An EMER example of multimodal emotion reasoning comparing Emotion-LLaMA with \ourmethod.}
\centering  
\scalebox{0.90}{
    \begin{tabular}{l|p{16cm} }
\toprule
\multicolumn{2}{l}{\bf EMER Example of Multimodal Emotion Reasoning} \\
\midrule
& {\includegraphics[height=3cm]{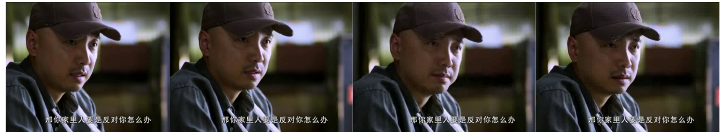}} \\
Subtitle (En) & {What will you do if your family opposes you?} \\
\midrule
Ground Truth & {In the video, we see a male wearing a baseball cap, with a focused expression, seemingly attentively listening to someone or contemplating a certain question. We can infer that the person in the video is likely seriously dealing with a problem or listening to someone speak, and his emotional state may be serious and focused. In the audio, it is impossible to determine the person's emotional state. In the text, the subtitle says, "What will you do if your family opposes you?" This sentence may be a question or concern from the male character towards someone. Based on the video clue of the male character's focused gaze and serious expression, as well as the audio clue of the inability to determine the person's emotional state, we can deduce that this sentence may carry a serious or worried tone. Therefore, this sentence may be the male character expressing concern or opposition towards someone's decision or contemplating a certain question.}  \\
\midrule
Emotion-LLaMA & {In the video, a man sits in a restaurant, looking directly at the camera, and speaks in a neutral tone his facial expression is neutral, with no visible emotional cues the tone of his voice is also neutral, as he asks a question about a hypothetical situation the lack of emotional cues and neutral tone suggest that the man is not experiencing any strong emotions, and his question appears to be a genuine inquiry rather than an emotional outburst.}  \\
\midrule
\ourmethod & {In the video, a man sits at a gas station, displaying a mix of facial expressions, including a lowered brow, tightened lips, and lips apart showing teeth, suggesting a sense of concern or anxiety. His tone and intonation are normal, but his words imply a sense of uncertainty and apprehension, indicating that he is worried about a potential problem or conflict.}  \\
\midrule
\end{tabular}
}
\vspace{5mm}
\label{tab:EMER_sample_comparison}
\end{minipage}
\end{table*}